\documentclass[11pt]{article}

 \usepackage[preprint]{acl}
\usepackage{times}
\usepackage{microtype}
\usepackage{subcaption}
\usepackage{graphicx}
\usepackage{amsmath}
\usepackage{amsfonts}
\usepackage{amssymb}
\usepackage{float}      
\usepackage{algorithm}  
\usepackage{algpseudocode}
\usepackage{multirow} 
\usepackage{booktabs} 
\usepackage{colortbl}
\usepackage{xurl}
\usepackage[dvipsnames]{xcolor}
\definecolor{darkgreen}{rgb}{0,0.5,0}
\definecolor{fedseblue}{HTML}{ECF4FF}
\usepackage{amsthm}
\usepackage{hhline}
\definecolor{lightlightgray}{gray}{0.95}

\newtheorem{lemma}{Lemma}
\newtheorem{theorem}{Theorem}

\usepackage[english]{babel}

\title{Fed-SE: Federated Self-Evolution for Cross-Environment Knowledge Transfer in Privacy-Constrained LLM Agents}

\author{
  Xiang Chen\textsuperscript{1} \quad
  Yuling Shi\textsuperscript{2} \quad
  Qizhen Lan\textsuperscript{3} \quad
  Yuchao Qiu\textsuperscript{1} \quad \\
  \textbf{Min Wang\textsuperscript{4} \quad
  Xiaodong Gu\textsuperscript{2} \quad
  Yanfu Yan\textsuperscript{1}} \\
  \textsuperscript{1}Zhejiang University \quad
  \textsuperscript{2}Shanghai Jiao Tong University \quad \\
  \textsuperscript{3}UTHealth Houston \quad
  \textsuperscript{4}University of Pennsylvania \\
  \texttt{\{chenxianghz,yanfu\}@zju.edu.cn}
}

\begin{document}

\maketitle

\begin{abstract}
LLM (Large Language Model) agents are widely deployed in complex interactive tasks, yet privacy constraints often preclude centralized optimization and co-evolution across dynamic environments. Despite the demonstrated success of Federated Learning (FL) on static datasets, its effectiveness in open-ended, self-evolving agent systems remains largely unexplored. In such settings, the direct application of standard FL is particularly challenging, as heterogeneous tasks and sparse, trajectory-level reward signals give rise to severe gradient instability, which undermines the global optimization process. To bridge this gap, we propose Fed-SE, a Federated Self-Evolution framework for LLM agents that establishes a local evolution–global aggregation paradigm. 
Locally, agents employ parameter-efficient fine-tuning on filtered, high-return trajectories to achieve stable gradient updates. Globally, Fed-SE aggregates updates within a low-rank subspace, reducing communication cost across clients. Experiments across five heterogeneous environments demonstrate that Fed-SE improves average task success rates by 10\% over the state-of-the-art FedIT, validating its effectiveness in cross-environment knowledge transfer under privacy constraints.\footnote{Our code is available at 
\url{https://github.com/Soever/Federated-Agents-Evolution}
}

\end{abstract}
\begin{figure}[ht]
  \centering
  \includegraphics[width=\columnwidth]{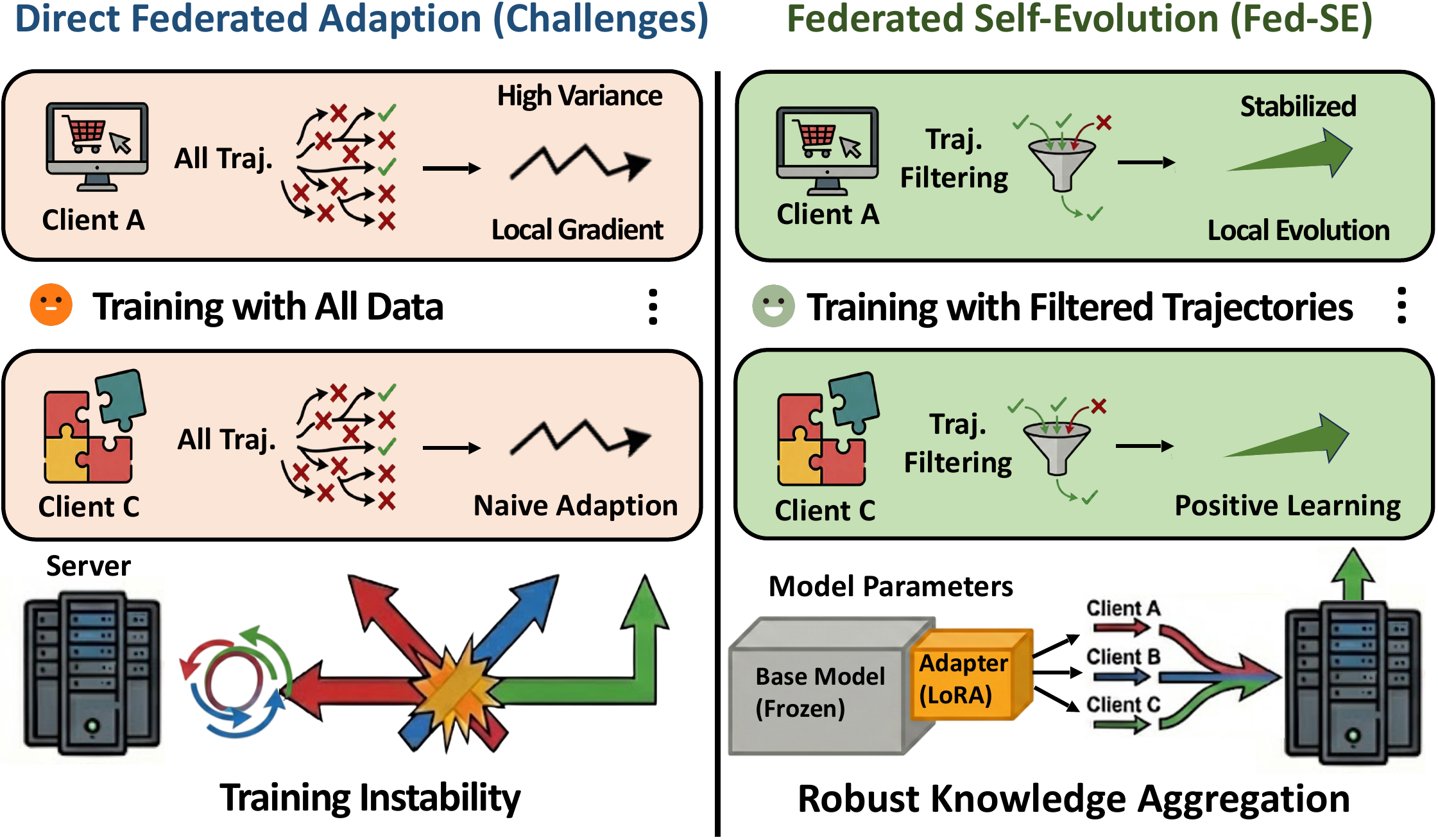}
  \caption{Static federated methods limit agent adaptation. Directly using FL suffers from high variance and gradient instability. Fed-SE addresses these issues via trajectory filtering for stable updates and low-rank aggregation for efficient communication.}
  \vspace{-0.25cm}
\end{figure}
\section{Introduction}

LLM-based agents have demonstrated significant potential in complex interactive tasks, ranging from embodied intelligence to online service systems~\citep{DBLP:conf/corl/ZitkovichYXXXXW23,DBLP:conf/rss/BelkhaleDXSVTCD24,li2025swe,peng2025swe,shi2024code,openai2024gpt4technicalreport,shi2025longcodezip,yang2025elaboration,zhang2025tri,zhang2025kabb,zhang2025mat}. The enhancement of agent capabilities typically relies on the accumulation of experience through continuous interaction with environments \citep{liu2025agent0vlexploringselfevolvingagent,chen2025swe,fang2025comprehensivesurveyselfevolvingai,cai2025flexcontinuousagentevolution}, which enables agents to internalize task-specific knowledge and refine decision logic, tool usage strategies, and long-horizon planning. 

However, as real-world deployments are often subject to privacy constraints, platform compliance requirements and business risk controls preclude the centralized aggregation of raw interaction data~\citep{yang2019federatedmachinelearningconcept,li2021federatedlearningnoniiddata,kairouz2021advancesopenproblemsfederated}. Consequently, agents operating across multiple environments are optimized in isolation without cross-environment knowledge sharing~\citep{cheng2024exploringlargelanguagemodel,Silagadze_2023}, limiting the development of generalizable agent capabilities~\citep{zala2024envgengeneratingadaptingenvironments,he2025advancinglanguagemultiagentlearning}.

While Federated Learning (FL)~\citep{FedAvg} offers a paradigm for collaborative training without exporting raw data, its application has largely been confined to static offline corpora~\citep{DBLP:conf/aaai/Jajoo25,DBLP:conf/kdd/WuLLD024}. Alternative approaches~\citep{shi2025privacyenhancingparadigmsfederatedmultiagent,wu2024federatedincontextllmagent} avoid parameter training with context exchange but struggle to consolidate interaction experience into generalizable model parameters. Therefore, the open-ended, online self-evolution of agents across distributed environments still remains largely underexplored.

Fundamental challenges arise when directly applying FL to online self-evolution of LLM agents. Specifically, Federated learning methods~\citep{FedAvg,zhang-etal-2023-fedpetuning} like FedIT~\citep{zhang2024buildingfederatedgptfederated} typically rely on clients producing locally meaningful and aggregatable updates from reasonably well-behaved training data. In cross-environment agent training, however, data are online-generated trajectories~\citep{
pmlr-v151-jin22a,hwang2025federatedreinforcementlearningheterogeneous}. Under sparse rewards, a large fraction of trajectories carry little learning signal, which inflates the variance of policy-gradient estimates and causes severe gradient instability~\citep{li2025shapingsparserewardsreinforcement,bjorck2022highvarianceunavoidablerl}, thereby destabilizing global aggregation. Furthermore, the parameter scale of LLMs renders full-parameter synchronization prohibitively expensive~\citep{qi2024fdlorapersonalizedfederatedlearning,singhal2025fedexloraexactaggregationfederated}.

To address these challenges, we propose Federated Self-Evolution (Fed-SE), a framework for cross-environment knowledge transfer under privacy constraints. Specifically, clients perform local optimization on filtered successful trajectories to stabilize gradients, while the server aggregates updates within a low-rank subspace to reduce communication cost. Extensive evaluations across five heterogeneous environments and five base models demonstrate the effectiveness of Fed-SE.
The main contributions are summarized as follows:
\begin{itemize}

\item We explore cross-environment knowledge transfer for LLM agents under privacy constraints and propose Fed-SE, the \textbf{first} framework addressing this problem.

\item Fed-SE combines success-trajectory filtering with experience accumulation for low-variance local optimization, and employs parameter-efficient aggregation within the low-rank adapter space for practical deployment.

\item Experiments on five environments show 10\% absolute improvements over FedIT, with larger gains on long-horizon tasks.

\end{itemize}

\section{Preliminaries and Problem Setup}

Agent-environment interaction is modeled as a  Partially Observable Markov Decision Process (POMDP). At timestep $t$, the policy $\pi_\theta$, parameterized by $\theta$, generates a reasoning chain $h_t$ and action $a_t$ based on instruction $u$ and history $\mathcal{H}_{t-1}$. The probability of a trajectory $\tau$ in environment $e$ with horizon $T$ decomposes as:
\begin{equation}
\begin{split}
    p_\theta(\tau \mid e, u) = &\prod_{t=1}^T \Big( \pi_\theta(h_t \mid u, \mathcal{H}_{t-1}, o_t) \\
    & \cdot \pi_\theta(a_t \mid u, \mathcal{H}_{t-1}, o_t, h_t) \Big),
\end{split}
\end{equation}
where $o_t$ denotes the observation received at step $t$, $\mathcal{H}_{t-1} = (o_1, h_1, a_1, \ldots, o_{t-1}, h_{t-1}, a_{t-1})$ denotes the interaction history.

Consider a federated system with $K$ clients, where each client $k$ holds a private environment $e_k$ characterized by unique transition dynamics $\mathcal{P}_k$ and task distributions $p_k(u)$. This heterogeneity induces non-identical local datasets $\mathcal{D}_k$:
\begin{equation}
\begin{split}
    \mathcal{D}_k = \{(\tau, R(\tau)) \mid \tau &\sim \pi_\theta(\cdot \mid e_k, u), \\
    u &\sim p_k(u)\},
\end{split}
\end{equation}
where $R(\tau) \in \{0, 1\}$ is the sparse binary reward indicating task success.

The global goal is to maximize the weighted sum of local expected returns $J_k(\theta) = \mathbb{E}_{\tau \sim \mathcal{D}_k} [R(\tau)]$ via decentralized updates, strictly prohibiting raw trajectory sharing:
\begin{equation}
    \max_\theta J(\theta) = \sum_{k=1}^K \omega_k J_k(\theta),
\end{equation}
where $\omega_k$ represents the aggregation weight for client $k$ ($\sum \omega_k = 1$). 

This setting introduces several challenges. Sparse binary rewards cause most trajectories to carry little learning signal, leading to gradient instability that destabilizes local updates. Heterogeneous environment dynamics further exacerbate this issue by inducing divergent gradient directions across clients. Moreover, the parameter scale of LLMs renders full-parameter synchronization prohibitively expensive.

\section{Methodology}
\begin{figure*}
    \centering
    \includegraphics[width=\textwidth]{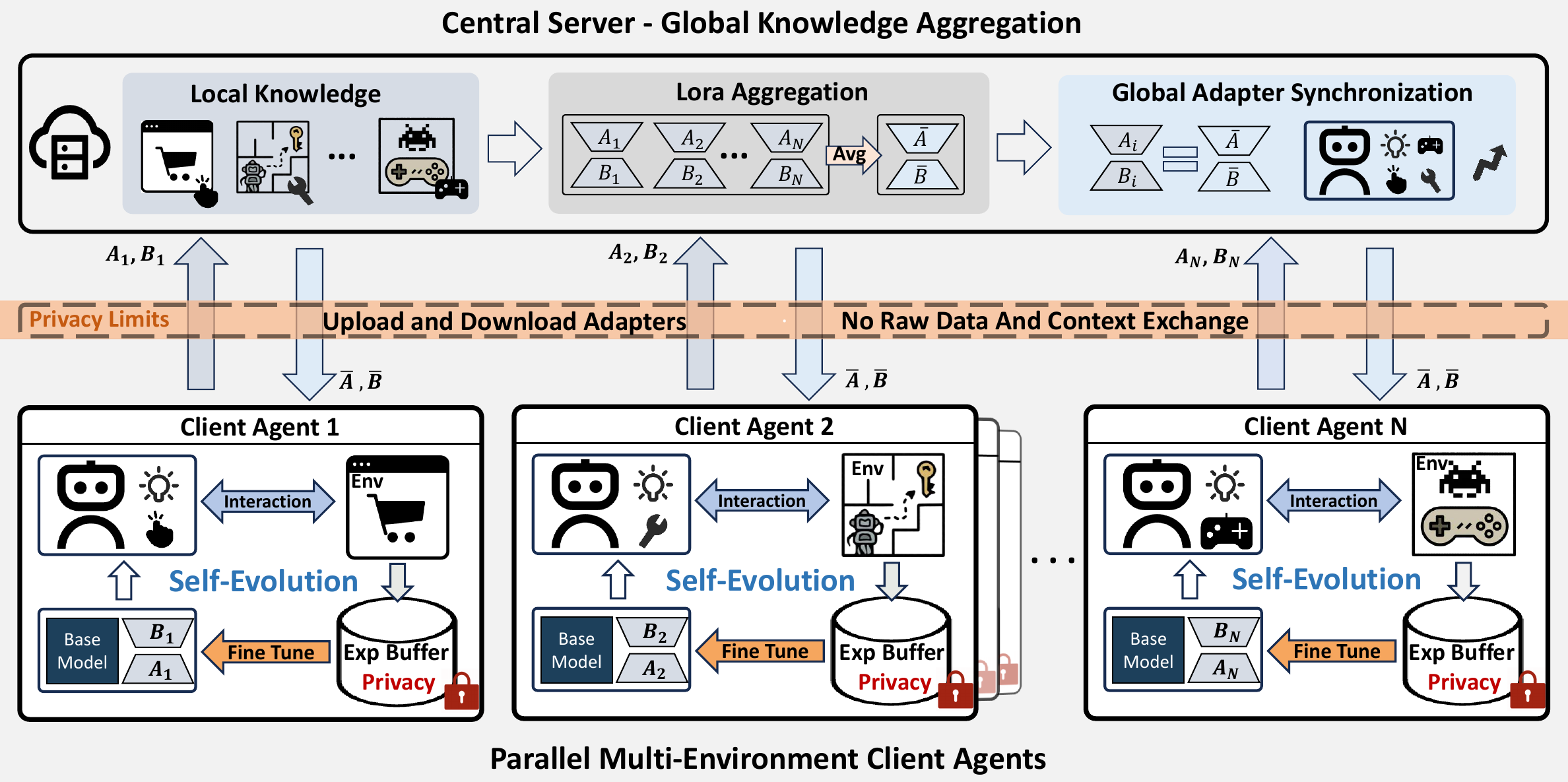}
    \caption{\textbf{Overview of the Fed-SE Framework.} Fed-SE operates in two phases: (1) local self-evolution, where clients optimize LoRA adapters using filtered successful trajectories, and (2) global aggregation, where the server averages distributed adapters and synchronizes them across clients.}
    \label{fig:framework}
\end{figure*}
The Federated Self-Evolution (Fed-SE) framework enables collaborative agent training under privacy constraints. As illustrated in Figure \ref{fig:framework}, the training operates iteratively: each round consists of Local Agent Self-Evolution (Section \ref{sec:local_evolution}), where clients optimize adapters using filtered trajectories, and Global Knowledge Aggregation (Section \ref{sec:global_aggregation}), where the server unifies distributed knowledge.

\subsection{Local Agent Self-Evolution}
\label{sec:local_evolution}
To mitigate the high gradient variance characteristic of sparse reward settings, the standard RL objective is optimized via a surrogate lower bound. By leveraging importance sampling and treating the policy from the previous iteration as the reference distribution, the maximization of expected return is theoretically approximated by performing Maximum Likelihood Estimation (MLE) solely on the distribution of successful trajectories $\mathcal{D}^+ $ (see Appendix for details). Consequently, the optimization problem is formulated as:
\begin{equation}
\max_\theta \mathbb{E}_{\tau \sim \mathcal{D}^+} [\log \pi_{\theta}(\tau)].
\end{equation}
Guided by this formulation, the local evolution process proceeds as follows.

\paragraph{Exploration and Filtering.}The agent first interacts with the local environment using the current policy to generate exploration trajectories. Based on the binary reward signal, the subset of successful trajectories $\mathcal{D}_{k,t}^{\text{succ}}$ is filtered out as follows:
\begin{equation}
\mathcal{D}_{k,t}^{\text{succ}} = \{ \tau \mid \tau \sim \pi_{\Theta, \phi_t}, R(\tau) = 1 \},
\end{equation}
where $\Theta$ represents the frozen base model parameters, and $\phi_t$ denotes the trainable adapter parameters at round $t$.

\paragraph{Experience Accumulation.}To alleviate distribution shift, a cumulative experience buffer merges historical data with newly discovered successful trajectories:
\begin{equation}
\mathcal{D}_{k,t}^{\text{train}} = \mathcal{D}_{k,t-1}^{\text{train}} \cup \mathcal{D}_{k,t}^{\text{succ}}.
\end{equation}
The buffer is initialized with a small set of expert demonstrations $\mathcal{D}_{k,0}^{\text{train}} = \mathcal{D}_{k}^{\text{expert}}$ to enable stable learning from the first round.
\paragraph{Parameter-Efficient Fine-Tuning.} The base model $\Theta$ is kept frozen, and only optimize the lightweight adapter parameters $\phi$ for efficient fine-tuning. Specifically, we minimize the negative log-probability loss on the accumulated dataset as follows:
\begin{equation}
\mathcal{L}(\phi) = -\mathbb{E}_{\tau \sim \mathcal{D}_{k,t}^{\text{train}}} \bigg[ \sum_{j=1}^{|\tau|} \ell_j(\phi) \bigg],
\end{equation}
where $\ell_j(\phi)$ denotes the joint log-likelihood of the generated reasoning chain and action at step $j$, defined as:
\begin{equation}
\begin{split}
\ell_j(\phi) = &\log \pi_{\Theta, \phi}(h_j \mid u, \mathcal{H}_{j-1}, o_j) \\
&+ \log \pi_{\Theta, \phi}(a_j \mid u, \mathcal{H}_{j-1}, o_j, h_j).
\end{split}
\end{equation}
where $o_j$ is the observation at step $j$, $h_j$ denotes the generated reasoning chain, and $a_j$ is the subsequent action.

\subsection{Global Knowledge Aggregation}
\label{sec:global_aggregation}

The global phase distills environment-specific experiences into generalizable capabilities under strict communication constraints.

\paragraph{Low-Rank Subspace Aggregation.} To reduce communication overhead and mitigate negative transfer from task heterogeneity, aggregation operates within the low-rank adapter space. The server computes unweighted averaging to prevent bias toward environments with abundant easy trajectories:
\begin{equation}
\overline{\phi}_{t} = \frac{1}{K} \sum_{k=1}^{K} \phi_{t,k}.
\end{equation}

\paragraph{Global Parameter Synchronization.} To mitigate client drift, local parameters are reset to the global consensus before each round:
\begin{equation}
\phi_{t+1,k}=\overline{\phi}_{t},\quad \forall k\in{1,\dots,K}.
\end{equation}

\begin{algorithm}[h]

\caption{Federated Self-Evolution}

\label{alg:fed_se}

\begin{algorithmic}[1]

\Require Base model $\Theta$, initial LoRA $\phi_0$, clients $\{1,\dots,K\}$, rounds $T$

\Ensure Optimized global LoRA parameters $\phi_T$

\For{$t = 0$ \textbf{to} $T-1$}
    \State Broadcast $\phi_t$ to all clients $k \in \{1,\dots,K\}$
    \ForAll{client $k$ \textbf{in parallel}}
        \State Explore with $\pi_{\Theta,\phi_t}$, filter successful trajectories
        \State $\mathcal{D}^{\text{train}}_{k,t} \gets \mathcal{D}^{\text{train}}_{k,t-1} \cup \{ \tau \mid \tau \sim \pi_{\Theta,\phi_t}, R(\tau)=1 \}$
        \State $\phi_{t+1,k} \gets \arg\min_{\phi} \mathcal{L}(\phi; \mathcal{D}^{\text{train}}_{k,t})$ \Comment{Init $\phi$ with $\phi_t$}
    \EndFor
    \State $\phi_{t+1} \gets \frac{1}{K} \sum_{k=1}^{K} \phi_{t+1,k}$ 

\EndFor

\end{algorithmic}

\end{algorithm}
 
\subsection{Theoretical Analysis}

\paragraph{Assumptions.} Standard assumptions in federated optimization are adopted: (1) $L$-smoothness of each local objective $f_k$, (2) unbiased stochastic gradients with variance bounded by $\sigma^2$, and (3) gradient norms bounded by $G$. 

\paragraph{Convergence Result.} Define the gradient heterogeneity at round $t$ as:
\begin{equation}
\zeta^2(\phi_t) = \frac{1}{K}\sum_{k=1}^K \|\nabla f_k(\phi_t) - \nabla f(\phi_t)\|^2,
\end{equation}
where $\nabla f_k(\phi_t)$ and $\nabla f(\phi_t)$ denote the local and global gradients at round $t$, respectively.

\begin{theorem}[Convergence of Fed-SE]
Under Assumptions (1)-(3), with a learning rate satisfying $\eta \le \frac{1}{4LE}$, the convergence bound after $T$ communication rounds is given by:
\begin{equation}
\begin{split}
\frac{1}{T}\sum_{t=0}^{T-1} \mathbb{E}\|\nabla f(\phi_t)\|^2 
&= O\!\left(\frac{1}{\eta ET}\right) + O\!\left(\frac{\eta\sigma^2}{K}\right) \\
&\quad + O\!\left(\eta E \bar{\zeta}^2\right),
\end{split}
\end{equation}
where $E$ is the number of local update steps per round, $\eta$ represents the learning rate, $K$ is the number of clients, and $\bar{\zeta}^2 = \frac{1}{T}\sum_{t}\zeta^2(\phi_t)$ represents the path-averaged heterogeneity.
\end{theorem}

Theorem 1 guarantees convergence to a stationary point of the global average objective. Complete proofs are provided in Appendix~\ref{sec:appendix_theory}.

\section{Experimental Results}

\subsection{Experiments Setting}

\begin{table*}[t]
\centering
\small
\begin{tabular}{clccccc|c}
\toprule
Base Model & Method & BabyAI & WebShop & TextCraft & Maze & Wordle & \textbf{Avg} \\
\midrule
\rowcolor{lightlightgray} \multicolumn{8}{c}{\textbf{\textit{\footnotesize Main Results}}} \\
\multirow{5}{*}{\textbf{Qwen2.5-7B}}
  & Pre-Trained  & 67.8 & 2.0  & 4.0  & 32.0 & \underline{20.0} & 18.6 \\
  & Local        & \underline{90.0} & 65.0 & \underline{63.0} & \underline{48.0} & 12.0 & 65.7 (\textcolor{darkgreen}{+47.0}) \\
  & Centralized  & 88.9 & \underline{71.0} & 59.0 & 40.0 & 8.0  & 66.6 (\textcolor{darkgreen}{+48.0}) \\
  & FedIT       & 83.3 & 67.5 & 54.0 & 28.0 & \underline{20.0} & 62.7 (\textcolor{darkgreen}{+44.1}) \\
  & \textbf{Fed-SE} & \textbf{93.3} & \textbf{73.0} & \textbf{67.0} & \textbf{68.0} & \textbf{32.0} & \cellcolor{fedseblue}\textbf{73.2} (\textbf{\textcolor{darkgreen}{+54.5}}) \\
\midrule
\rowcolor{lightlightgray} \multicolumn{8}{c}{\textit{\footnotesize Different Model Family}} \\
\multirow{5}{*}{Llama2-7B}
  & Pre-Trained  & 23.3 & 0.0  & 0.0  & 20.0 & 0.0  & 5.9 \\
  & Local        & 67.8 & 54.0 & 33.0 & 20.0 & \underline{4.0} & 47.3 (\textcolor{darkgreen}{+41.4}) \\
  & Centralized  & 61.1 & 59.0 & 36.0 & \underline{32.0} & 0.0  & 49.3 (\textcolor{darkgreen}{+43.4}) \\
  & FedIT       & \underline{70.0} & \underline{61.0} & \underline{49.0} & 28.0 & \textbf{16.0} & 55.7 (\textcolor{darkgreen}{+49.8}) \\
  & \textbf{Fed-SE} & \textbf{92.2} & \textbf{66.0} & \textbf{52.0} & \textbf{80.0} & \textbf{16.0} & \cellcolor{fedseblue}\textbf{66.1} (\textbf{\textcolor{darkgreen}{+60.2}}) \\
\midrule
\rowcolor{lightlightgray} \multicolumn{8}{c}{\textit{\footnotesize Different Model Scale}} \\
\multirow{5}{*}{Qwen2.5-3B}
  & Pre-Trained  & 48.9 & 3.0  & 1.0  & 24.0 & 4.0  & 13.2 \\
  & Local        & 81.1 & 60.0 & 46.0 & \underline{28.0} & \textbf{16.0} & 56.8 (\textcolor{darkgreen}{+43.6}) \\
  & Centralized  & \underline{83.3} & 61.5 & 48.0 & 20.0 & 4.0  & 57.3 (\textcolor{darkgreen}{+44.1}) \\
  & FedIT       & 72.2 & \underline{63.5} & \underline{54.0} & 20.0 & 4.0  & 57.3 (\textcolor{darkgreen}{+44.1}) \\
  & \textbf{Fed-SE} & \textbf{86.7} & \textbf{65.0} & \textbf{58.0} & \textbf{36.0} & \underline{8.0} & \cellcolor{fedseblue}\textbf{63.0} (\textbf{\textcolor{darkgreen}{+49.8}}) \\
\midrule
\rowcolor{lightlightgray} \multicolumn{8}{c}{\textit{\footnotesize Different Model Generation}} \\
\multirow{5}{*}{Qwen3-1.7B}
  & Pre-Trained  & 43.3 & 7.0  & 15.0 & \textbf{40.0} & 0.0  & 17.7 \\
  & Local        & \underline{73.3} & \underline{52.0} & 42.0 & \underline{32.0} & \textbf{4.0} & 50.2 (\textcolor{darkgreen}{+32.5}) \\
  & Centralized  & 70.0 & 44.5 & \textbf{54.0} & 0.0  & 0.0  & 46.8 (\textcolor{darkgreen}{+29.1}) \\
  & FedIT       & 61.1 & 50.5 & \underline{46.0} & 8.0  & \textbf{4.0} & 46.6 (\textcolor{darkgreen}{+28.9}) \\
  & \textbf{Fed-SE} & \textbf{74.4} & \textbf{57.5} & \underline{46.0} & \textbf{40.0} & \textbf{4.0} & \cellcolor{fedseblue}\textbf{54.3} (\textbf{\textcolor{darkgreen}{+36.6}}) \\

\specialrule{0.1pt}{2pt}{2pt}
\multirow{5}{*}{Qwen3-8B}
  & Pre-Trained  & 40.0 & 1.5  & 1.0  & 8.0  & \textbf{32.0} & 11.4 \\
  & Local        & \underline{85.6} & \underline{64.5} & \underline{70.0} & \underline{32.0} & \underline{12.0} & 65.2 (\textcolor{darkgreen}{+53.9}) \\
  & Centralized  & 78.9 & 33.5 & 63.0 & \underline{32.0} & 8.0  & 48.0 (\textcolor{darkgreen}{+36.6}) \\
  & FedIT       & \underline{85.6} & \underline{64.5} & 61.0 & 12.0 & 4.0  & 61.6 (\textcolor{darkgreen}{+50.2}) \\
  & \textbf{Fed-SE} & \textbf{92.2} & \textbf{68.0} & \textbf{71.0} & \textbf{64.0} & \underline{12.0} & \cellcolor{fedseblue}\textbf{70.2} (\textbf{\textcolor{darkgreen}{+58.9}}) \\
\bottomrule
\end{tabular}
\caption{Task-wise success rate across five base models. In the main results with Qwen2.5-7B, Fed-SE improves the average success rate by 10\% over FedIT. This advantage is consistently observed across different model families, scales, and generations.}
\label{tab:main_results}
\end{table*}
\begin{figure*}[t]
    \centering
    \includegraphics[width=\linewidth]{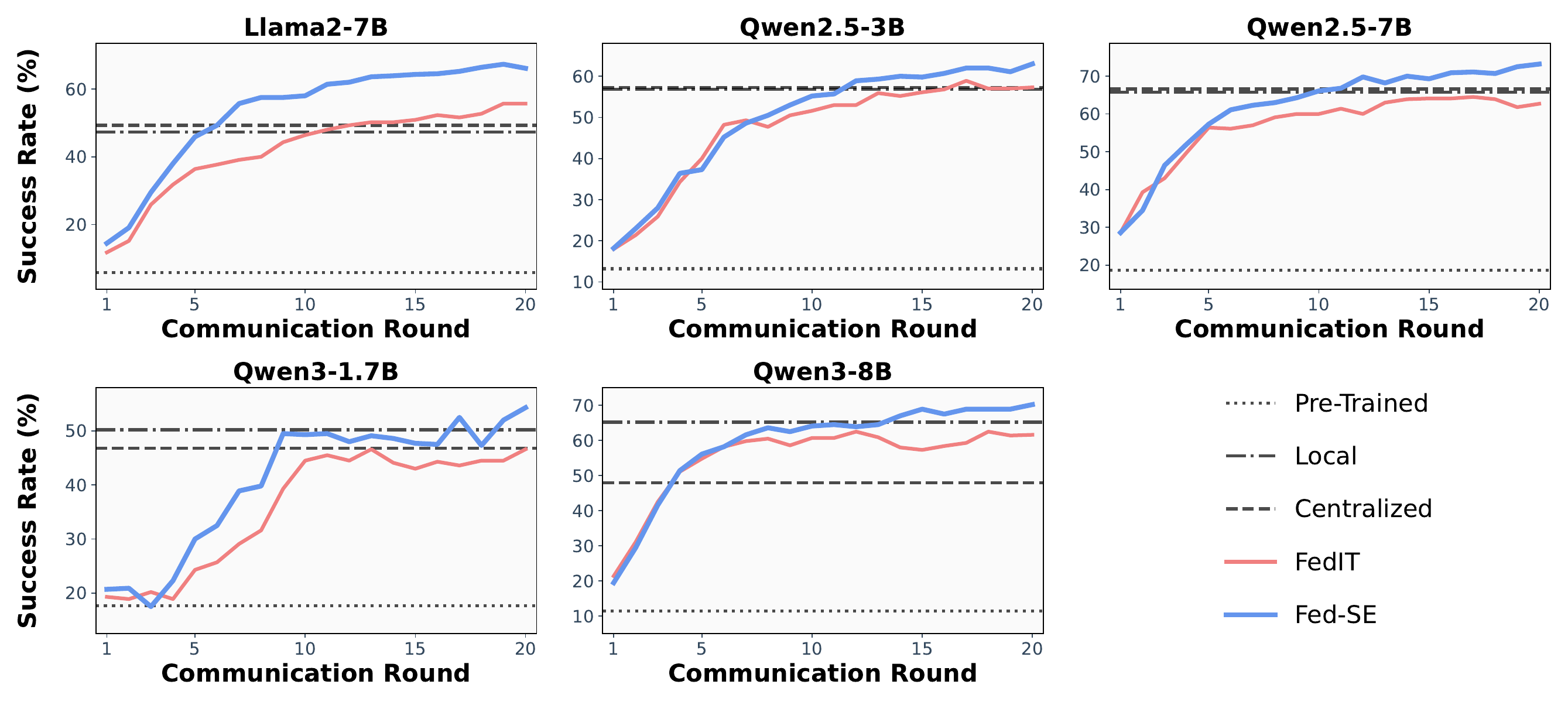}
    \caption{\textbf{Average success rate comparison across five base models.} The curves illustrate the training trajectories of Fed-SE and FedIT over communication rounds, while Local and Centralized represent static baselines trained on fixed datasets. Fed-SE achieves the highest average success rates after 20 communication rounds across diverse model architectures (Llama2, Qwen2.5, Qwen3) and parameter scales (1.7B to 8B).}
    \label{fig:main_results}
    \vspace{-0.5em}
\end{figure*}

\paragraph{Environment and Tasks.} 
Following the taxonomy established by \citet{lu2024fedhca2heteroclientfederatedmultitask}, 
federated multi-task learning scenarios can be categorized by 
the degree of task heterogeneity across clients. Our experimental setup adopts the most challenging configuration where each client is dedicated to a distinct environment, representing the extreme case of task heterogeneity. 

The proposed Fed-SE framework is evaluated across heterogeneous environments that cover diverse sequential decision-making capabilities: BabyAI~\citep{BabyAI} (embodied control and grounding), WebShop~\citep{webshop} (web interaction), TextCraft~\citep{textcraft} (hierarchical planning), MAZE~\citep{lmrl} (long-horizon memory), and Wordle~\citep{lmrl} (iterative reasoning). These environments have been widely adopted by prior work~\citep{liu2023agentbench,ma2024agentboardanalyticalevaluationboard,xi2024agentgym} for evaluating LLM-based agent capabilities, ensures a comprehensive assessment of agent generalization across distinct task dynamics. Performance is measured by task success rate, which indicates whether the agent successfully completes the given instruction within the environment.

\paragraph{Base Models.} 
Qwen2.5-7B \citep{Qwen2} is selected as the primary model for the main experiments. To comprehensively validate the generalizability of Fed-SE, the evaluation is also conducted across three dimensions: Llama2-7B-chat \citep{Llama2-7B-chat}  is included to examine whether Fed-SE's effectiveness transfers across different model families; Qwen2.5-3B \citep{Qwen2} is evaluated to investigate the impact of model scale within the same model family; Qwen3-1.7B and Qwen3-8B \citep{Qwen3} are incorporated to assess Fed-SE's applicability across different model architectures in the Qwen family.

\paragraph{Baselines.} As discussed in Section~\ref{sec:related}, to the best of our knowledge, no existing method addresses federated \textit{online self-evolution} for LLM agents. We establish a set of baselines to evaluate Fed-SE's performance gains over static and isolated training paradigms:
(1) \textbf{Pre-Trained}: The base model performs inference without fine-tuning.
(2) \textbf{Local}~\citep{hu2021loralowrankadaptationlarge}:  Agents fine-tuned independently on local static datasets using LoRA.
(3) \textbf{Centralized}~\citep{hu2021loralowrankadaptationlarge}: Static datasets aggregated for joint LoRA-based instruction tuning.
(4) \textbf{FedIT}~\citep{zhang2024buildingfederatedgptfederated}: FedIT represents a standard federated instruction-tuning approach based on \textit{static} expert demonstrations. We adapt FedIT,  rather than other related approaches~\citep{FedRLHF,wu2024federatedincontextllmagent}, from single-turn instruction–response pairs to multi-turn trajectory imitation for comparison, as this setting allows the contribution of online self-evolution to be evaluated more explicitly. 

Complete training implementation details are provided in Appendix~\ref{sec:impl_details}.

\subsection{Main Results}

As detailed in Table~\ref{tab:main_results} and Figure~\ref{fig:main_results}, Fed-SE achieves the highest average success rate across all base models, consistently outperforming all baselines. Through knowledge accumulation via local agent evolution and knowledge aggregation via global LoRA-based parameter sharing, Fed-SE enables cross-environment knowledge transfer under privacy constraints.

\paragraph{Knowledge Transfer under Privacy Constraints.} Table~\ref{tab:main_results} shows that compared to Fed-IT, which also operates under privacy constraints, Fed-SE achieves improvements across all five environments rather than on isolated tasks. On Qwen2.5-7B, Fed-SE outperforms Fed-IT on BabyAI (93.3\% vs. 83.3\%), WebShop (73.0\% vs. 67.5\%), TextCraft (67.0\% vs. 54.0\%), Maze (68.0\% vs. 28.0\%), and Wordle (32.0\% vs. 20.0\%). Similar patterns are observed across other base models. This indicates that combining online evolution with LoRA-based aggregation enables effective cross-environment knowledge transfer without exposing raw trajectories.

\vspace{-0.3em}

\paragraph{Knowledge Accumulation via Online Evolution.} Figure~\ref{fig:main_results} illustrates the training curves across communication rounds. The performance gap between Fed-SE and Fed-IT widens progressively as training proceeds, indicating that online evolution effectively converts interaction experiences into intrinsic model capabilities. However, this mechanism relies on the availability of successful trajectories. On Wordle, where successful experiences are difficult to obtain in early stages, the improvement remains limited.

\paragraph{Generalization Across Base Models.} As detailed in Table~\ref{tab:main_results} and Figure~\ref{fig:main_results}, Fed-SE maintains consistent improvements across different model families, scales, and generations, demonstrating the stability and robustness of the framework.

\section{Analysis} 
\subsection{Ablation Study}
\label{sec:ablation}
\begin{figure*}[!ht] 
    \centering
    \includegraphics[width=1\linewidth]{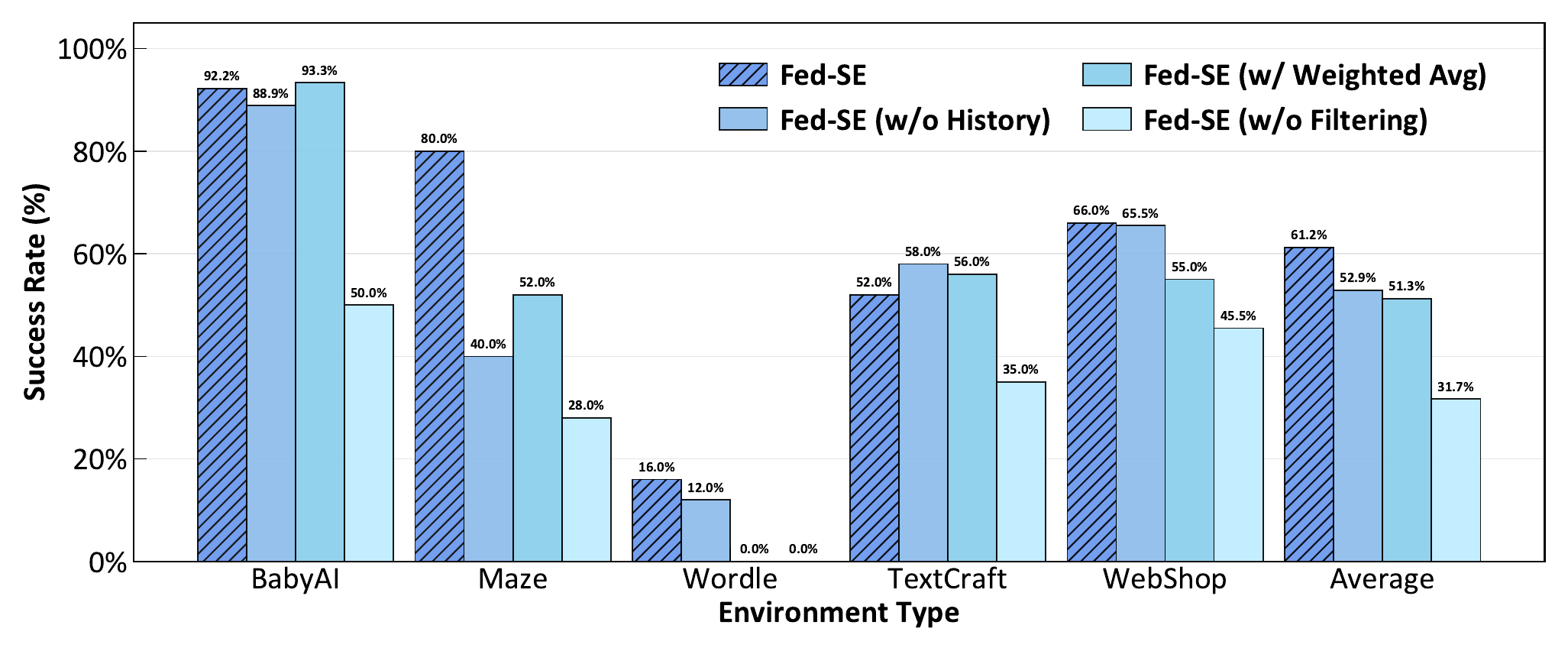}
    \caption{\textbf{Impact of Key Components on Final Performance.} Removing the success filter causes a catastrophic performance drop (-26\%), while excluding history or using weighted averaging also degrades the robust baseline (66\%).}
    \label{fig:ablation_bar}
\end{figure*}

To verify the effectiveness of key components in Fed-SE, we conduct ablation studies using Llama2-7B-Chat as the base model and construct three variants: (1) \textit{w/o History:} Removes the experience accumulation mechanism, fine-tuning only with new data from the current round; (2) \textit{w/o Filtering:} Removes the success filter, including failed trajectories in training; (3) \textit{w/ Weighted Avg:} Uses weighted averaging based on the number of successful trajectories during aggregation.

\begin{figure}[!ht] 
    \centering
    
    \begin{subfigure}[b]{\columnwidth} 
        \centering
       \includegraphics[width=\columnwidth]{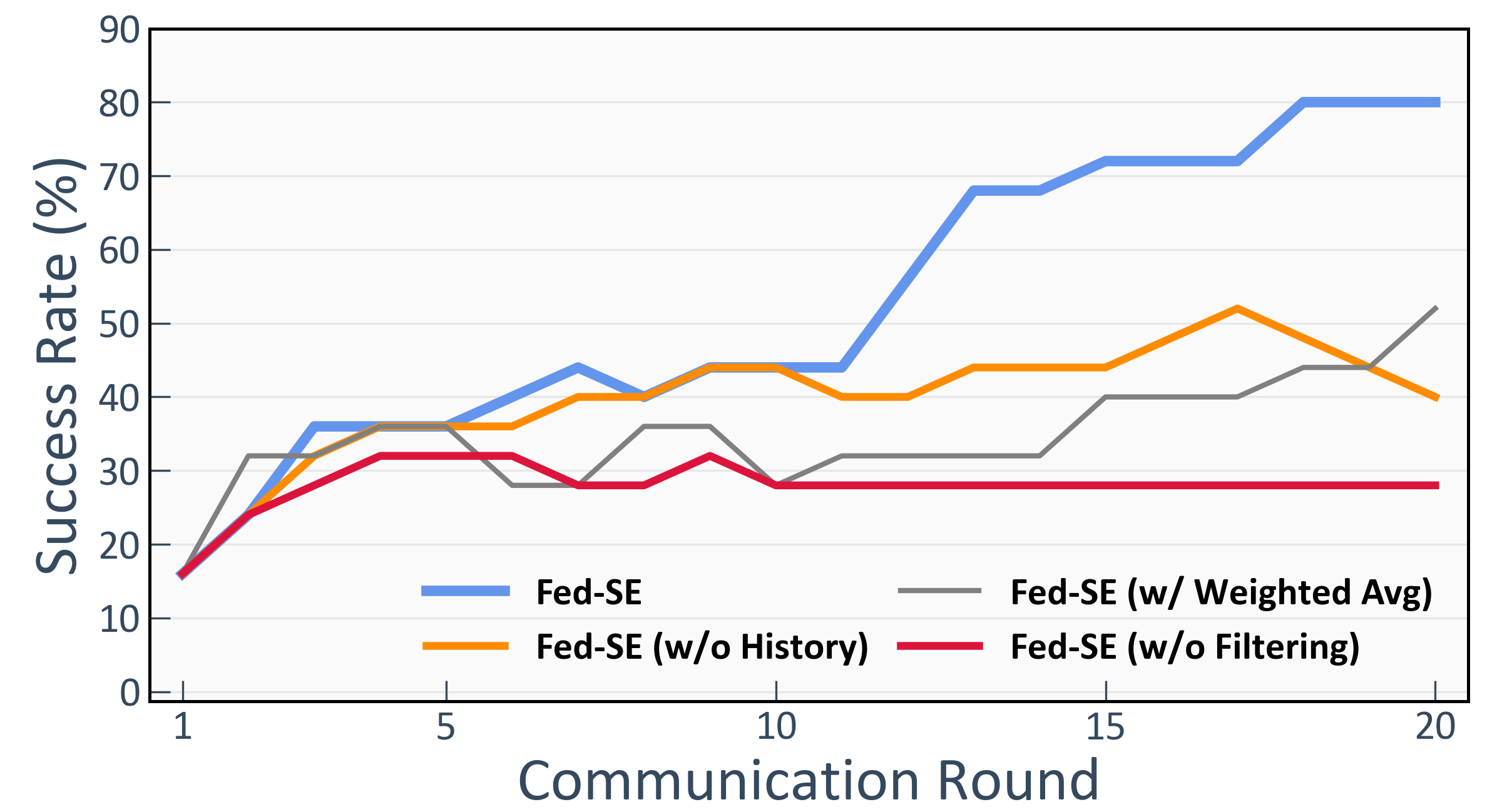}
        \caption{Performance Evolution on Maze}
        \label{fig:ablation_maze}
    \end{subfigure}

    \begin{subfigure}[b]{\columnwidth}
        \centering
        \includegraphics[width=\columnwidth]{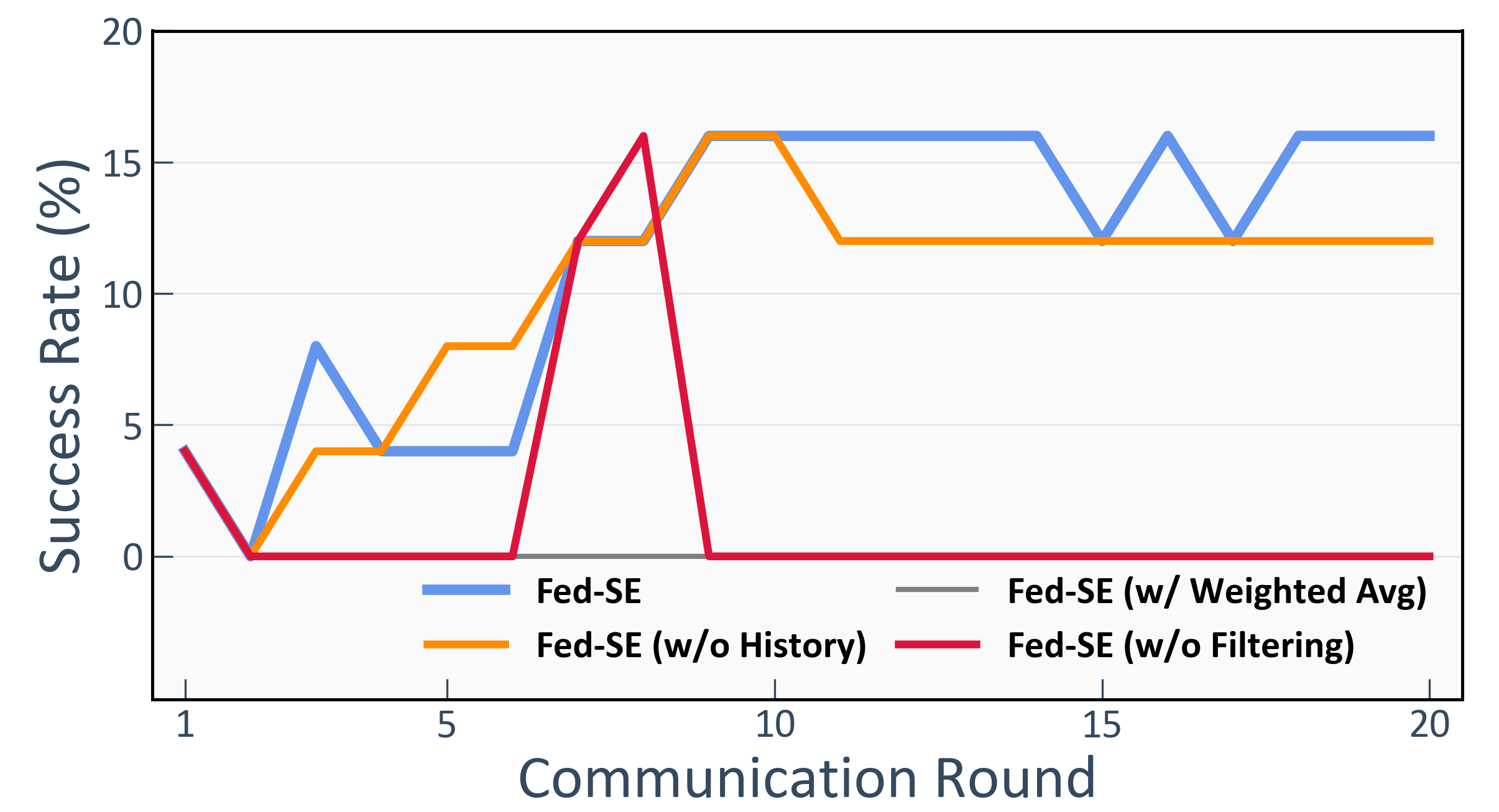}
        \caption{Performance Evolution on Wordle}
        \label{fig:ablation_wordle}
    \end{subfigure}
    
    \caption{\textbf{Evolution Process Analysis.} (a) The \textbf{Maze} task shows that removing history accumulation (w/o History) leads to suboptimal convergence. (b) The \textbf{Wordle} task demonstrates that removing the success filter (w/o Filtering) causes catastrophic performance collapse due to noise injection.}

    \label{fig:ablation_combined}
    \vspace{-0.5em}
\end{figure}

\paragraph{Overall Performance Analysis.} As shown in Figure \ref{fig:ablation_bar}, the full Fed-SE achieves the highest average success rate (66.1\%), outperforming all ablation variants on average. Notably, removing the success filter causes the most severe performance decay, with the average success rate plummeting to 40.5\%, while removing history experience and changing the aggregation strategy also lead to varying degrees of performance decline.

\paragraph{Importance of Cumulative History.} 
While w/o History (64.1\%) achieves a competitive average, this masks critical deficiencies in complex long-horizon tasks. Specifically, when evaluated on Maze (Figure \ref{fig:ablation_maze}), performance plateaus at 40.0\%, far below Fed-SE's 80.0\%. This significant drop indicates that relying solely on fresh data leads to the loss of prior capabilities during distribution adaptation. The historical buffer acts as an Experience Replay mechanism, effectively suppressing Catastrophic Forgetting and stabilizing policy oscillation against online distribution shifts, thereby sustaining the evolution of long-horizon planning.

\vspace{-1em}

\paragraph{Necessity of Success Filtering.} Removing the success filter leads to the most drastic decay. In Wordle (Figure \ref{fig:ablation_wordle}), success rates collapse to zero after round 8. This confirms that failed trajectories act as misleading signals under behavioral cloning assumptions. Without strict filtering, the model erroneously imitates failure, rapidly contaminating global parameters and diverting policy optimization from the true objective.

\paragraph{Robustness of Aggregation Strategy.} Weighted averaging (59.8\%) underperforms simple averaging, suggesting that under highly heterogeneous distributions, trajectory quantity does not correlate with gradient quality (e.g., BabyAI generates abundant simple samples). Weighted aggregation risks biasing the global model toward simpler tasks, weakening generalization on difficult ones. Therefore, simple averaging demonstrates superior robustness in balancing multi-task heterogeneity.

\subsection{Communication Efficiency}
Communication overhead in FL is a critical constraint for deployment. The Fed-SE framework achieves parameter-efficient transmission through LoRA adapters, making communication overhead linearly related to the rank ($r$). We analyze this trade-off using Llama2-7B-Chat as the base model.

\begin{figure}[!ht]
    \centering
    \includegraphics[width=\columnwidth]{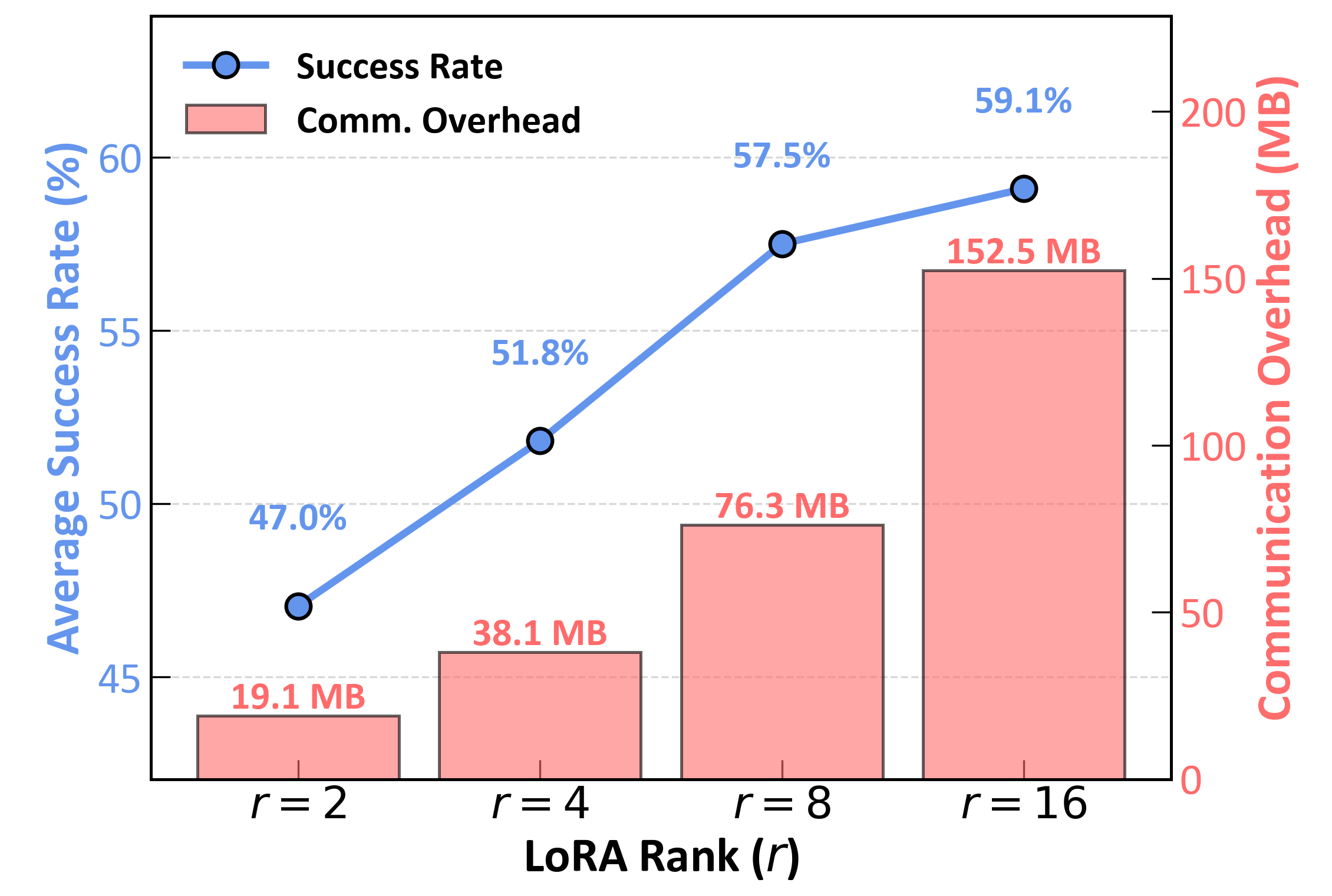}
    \caption{Trade-off between model performance and communication cost across different LoRA ranks.}
    \label{fig:rank_analysis}
\end{figure}

\paragraph{Trade-off between Performance and Cost.} Figure \ref{fig:rank_analysis} illustrates the trade-off between performance and communication cost under different ranks. The analysis reveals a typical trend of Diminishing Returns: increasing $r$ from 4 to 8 significantly boosts the success rate from $51.8\%$ to $57.5\%$ ($\mathbf{+5.7\%}$), indicating insufficient model capacity at $r=4$. However, further increasing $r$ from 8 to 16 yields only a marginal gain from $57.5\%$ to $59.1\%$ ($\mathbf{+1.6\%}$), while doubling the communication overhead ($\mathbf{76.3 \text{ MB} \to 152.5 \text{ MB}}$). This comparison compellingly demonstrates that $r=8$ represents the Optimal Trade-off Point between performance and communication cost.

\paragraph{Deployment Feasibility} As agents evolve, successful trajectories lengthen, leading to higher dynamic memory demands. Larger ranks (e.g., $r=16$) yield negligible gains but consume static memory needed for processing these long sequences. Thus, $r=8$ balances model capacity with memory constraints, preventing out-of-memory errors during later evolutionary stages.

\section{Related Work}
\label{sec:related}

\subsection{LLM Agents and Self-Evolution}
LLM agents have evolved from prompting-based systems to autonomous entities capable of planning, tool use, and multi-step reasoning~\citep{yao2023react,schick2023toolformer,wang2025vrag,wang2025vidorag,zhang2025gam,zhang2025mmcot}. Recent work enables agents to improve autonomously through self-evolution~\citep{tao2024survey, gao2025survey}, with self-improvement training emerging as the dominant paradigm, including self-rewarding~\citep{yuan2025selfrewardinglanguagemodels}, self-play~\citep{chen2024selfplayft}, and self-refinement~\citep{madaan2023selfrefine}. AgentGym~\citep{xi2024agentgym} integrates these mechanisms with diverse environments for continuous improvement. Memory mechanisms further support persistent learning, from reflective memory~\citep{Reflexion} to autonomous experience extraction~\citep{zhao2024expel} and continual learning via causal abstractions~\citep{majumder2023clin}. For cross-environment generalization, modular architectures~\citep{yin2024agentlumos}, self-evolving curricula~\citep{qi2025webrl}, and cross-task experience sharing~\citep{yang2024cops} show promising results. However, existing approaches universally assume centralized trajectory access, which is problematic for distributed, privacy-constrained  deployments.

\subsection{Federated Learning for Large Language Models and Agents}
Federated learning for LLMs has focused on parameter-efficient fine-tuning (PEFT) to mitigate high communication costs~\citep{FedEx-LoRA,FFA-Lora,LoraA2,FedCoT}. FedIT~\citep{zhang2024buildingfederatedgptfederated} and FedPETuning~\citep{zhang-etal-2023-fedpetuning} established the efficacy of federated instruction tuning. 
While FedRLHF~\citep{FedRLHF} establishes a framework for privacy-preserving policy optimization via client-specific feedback, it remains limited to single-turn preference alignment, lacking the multi-step reasoning capabilities required for autonomous agents. Meanwhile, FICAL~\citep{wu2024federatedincontextllmagent} pioneers federated in-context agent learning by transmitting knowledge compendiums, yet it relies on frozen models without parameter updates. Consequently, there is a lack of frameworks for the continuous, autonomous evolution of agents across heterogeneous environments under privacy constraints. To bridge this gap, we propose Fed-SE, a federated self-evolution framework that enables agents to improve their capabilities across distributed environments without raw data sharing. We adapt FedIT for comparison with our approach, as it performs federated parameter updates using static instruction data, thereby providing a clear contrast that isolates the contribution of online, trajectory-based self-evolution.

\section{Conclusion} 
This paper introduces Fed-SE, the first framework addressing federated online self-evolution for LLM agents, enabling cross-environment knowledge transfer under privacy constraints. Fed-SE establishes a local evolution–global aggregation paradigm where success-trajectory filtering with experience accumulation stabilizes local gradient updates, while unweighted low-rank aggregation reduces communication overhead and balances contributions across heterogeneous environments.
Experiments across five environments and five base models demonstrate 6–10\% absolute improvements over FedIT, with larger gains on long-horizon tasks. These results validate the feasibility of distilling generalizable agent capabilities from decentralized interaction experiences without raw data sharing.

\section*{Limitations}
While transmitting adapter parameters prevents raw data exposure, the framework does not currently incorporate cryptographic techniques such as Differential Privacy or Homomorphic Encryption. This design choice prioritizes the high parameter precision required for complex reasoning but may leave the system vulnerable to advanced gradient reconstruction attacks. Additionally, the current reliance on synchronous Federated Averaging assumes consistent client connectivity. In real-world edge deployments, device heterogeneity and network instability could induce straggler effects, potentially hindering convergence efficiency. Furthermore, the global aggregation mechanism relies on standard element-wise averaging, advanced aggregation strategies for agents will be explored in future work.

\section*{Ethics Considerations}
This work focuses on methodological contributions in simulated environments. Deploying self-evolving agents in real-world scenarios would require additional safety measures to ensure alignment with human values.

\bibliography{custom}
\clearpage

\appendix

\appendix
\section{Complete Proof of Theorem 1}
\label{sec:appendix_theory}

\subsection{Surrogate Objective Derivation}
\label{sec:surrogate_derivation}

The standard RL objective is $J_k(\phi) = \mathbb{E}_{\tau \sim \pi_\phi}[R(\tau)]$. Using importance sampling with a reference policy $\pi_{\text{ref}}$ (e.g., from the previous iteration):
\begin{equation}
J_k(\phi) = \mathbb{E}_{\tau \sim \pi_{\text{ref}}}\left[\frac{\pi_\phi(\tau)}{\pi_{\text{ref}}(\tau)} R(\tau)\right].
\end{equation}
For binary rewards $R(\tau) \in \{0,1\}$, maximizing $J_k(\phi)$ is equivalent to maximizing the likelihood ratio on successful trajectories $\mathcal{D}_k^+$. Applying Jensen's inequality to the logarithm:
\begin{equation}
\begin{split}
\log \mathbb{E}_{\tau \sim \mathcal{D}_k^+}\left[\frac{\pi_\phi(\tau)}{\pi_{\text{ref}}(\tau)}\right] &\ge \mathbb{E}_{\tau \sim \mathcal{D}_k^+}\left[\log \frac{\pi_\phi(\tau)}{\pi_{\text{ref}}(\tau)}\right] \\
&= \mathbb{E}_{\tau \sim \mathcal{D}_k^+}[\log \pi_\phi(\tau)] - C,
\end{split}
\end{equation}
where $C$ is a constant independent of $\phi$. This yields the surrogate optimization objective $f_k(\phi)$.

\subsection{Lemma: Client Drift Bound}

\begin{lemma}
\label{lemma:drift}
The parameter drift for client $k$ after $e$ local steps is bounded by:
\begin{equation}
\mathbb{E}\|\phi_{k,t}^{(e)} - \phi_t\|^2 \le e^2\eta^2(G^2 + \sigma^2).
\end{equation}
\end{lemma}
\begin{proof}
Using the update rule $\phi_{k,t}^{(e)} - \phi_t = -\eta \sum_{j=0}^{e-1} g_{k,t}^{(j)}$ and Cauchy-Schwarz inequality:
\begin{equation}
\begin{split}
\mathbb{E}\|\phi_{k,t}^{(e)} - \phi_t\|^2 &= \eta^2 \mathbb{E}\left\|\sum_{j=0}^{e-1} g_{k,t}^{(j)}\right\|^2 \\
&\le \eta^2 e \sum_{j=0}^{e-1} \mathbb{E}\|g_{k,t}^{(j)}\|^2.
\end{split}
\end{equation}
Since $\mathbb{E}\|g\|^2 \le G^2 + \sigma^2$ (Assumptions 2 and 3), the result follows.
\end{proof}

\subsection{Proof of Theorem 1}

\paragraph{Step 1: Smoothness Analysis.}
By the $L$-smoothness of the global objective $f$:
\begin{equation}
\begin{split}
\mathbb{E}[f(\phi_{t+1})] \le f(\phi_t) &+ \mathbb{E}\langle \nabla f(\phi_t), \Delta_t \rangle \\
&+ \frac{L}{2}\mathbb{E}\|\Delta_t\|^2,
\end{split}
\tag{A.1}
\end{equation}
where $\Delta_t = \phi_{t+1} - \phi_t = -\frac{\eta}{K}\sum_{k=1}^K \sum_{e=0}^{E-1} g_{k,t}^{(e)}$ is the global parameter update.

\paragraph{Step 2: Bounding the Inner Product.}
The inner product term is decomposed as:
\begin{equation}
\begin{split}
\mathbb{E}\langle \nabla f(\phi_t), \Delta_t \rangle &= -\frac{\eta}{K}\sum_{k,e} \mathbb{E}\langle \nabla f(\phi_t), \nabla f_k(\phi_{k,t}^{(e)}) \rangle \\
&= \underbrace{-\eta E \|\nabla f(\phi_t)\|^2}_{T_1} \\
&\quad \underbrace{-\frac{\eta}{K}\sum_{k,e} \langle \nabla f(\phi_t), \delta_{k,t}^{(e)} \rangle}_{T_2},
\end{split}
\end{equation}
where $\delta_{k,t}^{(e)} = \nabla f_k(\phi_{k,t}^{(e)}) - \nabla f_k(\phi_t)$.
Using Young's inequality regarding $T_2$:
\begin{equation}
\begin{split}
\langle \nabla f(\phi_t), \delta_{k,t}^{(e)} \rangle &\ge -\frac{1}{2}\|\nabla f(\phi_t)\|^2 - \frac{1}{2}\|\delta_{k,t}^{(e)}\|^2 \\
&\ge -\frac{1}{2}\|\nabla f(\phi_t)\|^2 - \frac{L^2}{2}\|\phi_{k,t}^{(e)} - \phi_t\|^2.
\end{split}
\end{equation}
Summing over $k$ and $e$, and defining the average drift $\mathcal{E}_t$:
\begin{equation}
T_2 \le \frac{\eta E}{2}\|\nabla f(\phi_t)\|^2 + \frac{\eta L^2 K E}{2} \mathcal{E}_t.
\end{equation}
Combining $T_1$ and $T_2$:
\begin{equation}
\mathbb{E}\langle \nabla f(\phi_t), \Delta_t \rangle \le -\frac{\eta E}{2}\|\nabla f(\phi_t)\|^2 + \frac{\eta L^2 K E}{2}\mathcal{E}_t.
\tag{A.2}
\end{equation}

\paragraph{Step 3: Bounding the Update Norm.}
The expected squared norm of the update $\mathbb{E}\|\Delta_t\|^2$ is bounded by:
\begin{equation}
\begin{split}
\mathbb{E}\|\Delta_t\|^2 &\le \frac{\eta^2 E \sigma^2}{K} + 3\eta^2 E^2 \Big( \|\nabla f(\phi_t)\|^2 \\
&\quad + \zeta^2(\phi_t) + L^2 \mathcal{E}_t \Big).
\end{split}
\tag{A.3}
\end{equation}

\paragraph{Step 4: Combining Results.}
Substituting (A.2) and (A.3) into (A.1):
\begin{equation}
\begin{split}
\mathbb{E}[f(\phi_{t+1})] \le f(\phi_t) &- \frac{\eta E}{2}(1 - 3\eta LE)\|\nabla f(\phi_t)\|^2 \\
&+ \frac{\eta^2 E L \sigma^2}{2K} + \frac{3\eta^2 E^2 L}{2}\zeta^2(\phi_t) \\
&+ \frac{L^2 \eta K E}{2}(1 + 3\eta LE)\mathcal{E}_t.
\end{split}
\end{equation}
With $\eta \le \frac{1}{4LE}$, we have $(1-3\eta LE) \ge \frac{1}{4}$. Substituting the drift bound from Lemma \ref{lemma:drift}:
\begin{equation}
\begin{split}
\frac{\eta E}{8}\|\nabla f(\phi_t)\|^2 \le &f(\phi_t) - \mathbb{E}[f(\phi_{t+1})] + \frac{\eta^2 E L \sigma^2}{2K} \\
&+ \frac{3\eta^2 E^2 L}{2}\zeta^2(\phi_t) + O(\eta^3).
\end{split}
\end{equation}

\paragraph{Step 5: Final Convergence Bound.}
Summing over $t=0$ to $T-1$ and rearranging:
\begin{equation}
\begin{split}
\frac{1}{T}\sum_{t=0}^{T-1} \mathbb{E}\|\nabla f(\phi_t)\|^2 \le &\frac{8(f_0 - f^*)}{\eta ET} + \frac{4\eta L\sigma^2}{K} \\
&+ 12\eta LE \bar{\zeta}^2 + O(\eta^2).
\end{split}
\end{equation}
This confirms the convergence rate stated in Theorem 1. 

\section{Implementation Details}
\label{sec:impl_details}
\begin{table*}[htbp]
\caption{Prompts for WebShop.}
\label{tab:prompt-webshop}
\centering
\small
\rule{\textwidth}{0.4pt}
\begin{minipage}{0.97\textwidth}
\vspace{0.5em}
\textbf{System Prompt:}\\[0.3em]
You are web shopping.
I will give you instructions about what to do.
You have to follow the instructions.
Every round I will give you an observation and a list of available actions, you have to respond an action based on the state and instruction.
You can use search action if search is available.
You can click one of the buttons in clickables.
An action should be of the following structure:
search[keywords]
click[value]
If the action is not valid, perform nothing.
Keywords in search are up to you, but the value in click must be a value in the list of available actions.
Remember that your keywords in search should be carefully designed.
Your response should use the following format:

Thought:
I think ...

Action:
click[something]
\vspace{0.5em}
\end{minipage}
\rule{\textwidth}{0.4pt}
\end{table*}

\begin{table*}[htbp]
\caption{Prompts for LMRL-Wordle.}
\label{tab:prompt-lmrlwordle}
\centering
\small
\rule{\textwidth}{0.4pt}
\begin{minipage}{0.97\textwidth}
\vspace{0.5em}
\textbf{Prompt:}\\[0.3em]
You are an expert wordle player. Your objective is to guess a hidden 5 letter word. You have 6 attempts to guess it correctly and you should try to guess it in as few attempts as possible. When guessing the word, you should format your word as a space separated sequence of letters, like "s h i r e" for example. After guessing the word, you will receive feedback from the game environment in the form of a sequence of 5 space separated letters like "b y g g b", where each letter indicates some information about the hidden word. The environment will return one of three letters - "b", "g", or "y" - for each letter in the word you guessed. We describe the meaning of each letter below:

"b": If the environment returns a "b", it means that the letter at that position in your guessed word is not in the hidden word.
"y": If the environment returns a "y", it means that the letter at that position in your guessed word is in the hidden word but is not in the correct position.
"g": If the environment returns a "g", it means that the letter at that position in your guessed word is in the hidden word and is in the correct position.

As a note, if you guess an invalid word (e.g. not a 5 letter word or a word not in the vocabulary), the environment will respond with an "invalid word" message. In general though, you should use this information returned by the environment to update your belief about what the hidden word might be and adjust your next guess accordingly.

Here is the complete list of valid vocabulary words that are accepted by the game:
```
\{\{vocab\}\}
```

Here is an example. If the current status of the game is given as:
```
guess 1: p a n i c
feedback 1: b b y b b
guess 2: f e l o n
feedback 2: g b b y g
```
Based on the feedback from the environment, you know that the first letter is "f", the last letter is "n", and there is an "o" somewhere in the word, but it is not in the second to last position. You also know that there is not a "p", "a", "i", "c", "e", or "l" in the word. Knowing this, you might guess the next word to be:
Thought:
I know that the first letter is "f", the last letter is "n", and there is an "o" somewhere in the word, but it is not in the second to last position. I also know that there is not a "p", "a", "i", "c", "e", or "l" in the word. A good word from the vocabulary to try might therefore be "f r o w n", since it is in the vocabulary, meets all known letter constraints, and we get to gain more information about the position of "o". Therefore this is a good guess to try next.

Action:
f r o w n

Formally, your return should be in this format:
Thought:
\textless{}Your Thought\textgreater{}

Action:
\textless{}The Word You Guess\textgreater{}

The guessed word is in the vocabulary, meets all known letter constraints, and we get to gain more information about the position of "o", so it is a good guess to try next.

Now let us start a new game. Remember, the word you guess should be strictly in the vocabulary. You should return your thought and your word strictly in the formation mentioned above.
\vspace{0.5em}
\end{minipage}
\rule{\textwidth}{0.4pt}
\end{table*}
\begin{table*}[htbp]
\caption{Prompts for BabyAI.}
\label{tab:prompt-babyai}
\centering
\small
\rule{\textwidth}{0.4pt}
\begin{minipage}{0.97\textwidth}
\vspace{0.5em}
\textbf{Prompt:}\\[0.3em]
You are an exploration master that wants to finish every goal you are given. Every round I will give you an observation, and you have to respond an action and your thought based on the observation to finish the given task. You are placed in a room and you need to accomplish the given goal with actions.

You can use the following actions:

- turn right

- turn left

- move forward

- go to \textless{}obj\textgreater{} \textless{}id\textgreater{}

- pick up \textless{}obj\textgreater{} \textless{}id\textgreater{}

- go through \textless{}door\textgreater{} \textless{}id\textgreater{}: \textless{}door\textgreater{} must be an open door.

- toggle and go through \textless{}door\textgreater{} \textless{}id\textgreater{}: \textless{}door\textgreater{} can be a closed door or a locked door. If you want to open a locked door, you need to carry a key that is of the same color as the locked door.

- toggle: there is a closed or locked door right in front of you and you can toggle it.
Your response should use the following format:
Thought:
\textless{}Your Thought\textgreater{}

Action:
\textless{}Your Action\textgreater{}
\vspace{0.5em}
\end{minipage}
\rule{\textwidth}{0.4pt}
\end{table*}

\begin{table*}[htbp]
\caption{Prompts for TextCraft.}
\label{tab:prompt-textcraft}
\centering
\small
\rule{\textwidth}{0.4pt}
\begin{minipage}{0.97\textwidth}
\vspace{0.5em}
\textbf{Prompt:}\\[0.3em]
You are given few useful crafting recipes to craft items in Minecraft. Crafting commands are of the format "craft [target object] using [input ingredients]".
Every round I will give you an observation, you have to respond an action based on the state and instruction. You can "get" an object (ingredients) from the inventory or the environment, look-up the game inventory by "inventory", or "craft" (target) using any of the crafting commands. You can use ONLY these crafting commands provided, do not use your own crafting commands. However, if the crafting command uses a generic ingredient like "planks", you can use special types of the same ingredient e.g. "dark oak planks" in the command instead.
Your response should use the following format:

Thought:
 ...

Action:
 ...
\vspace{0.5em}
\end{minipage}
\rule{\textwidth}{0.4pt}
\end{table*}

\begin{table*}[htbp]
\caption{Prompts for LMRL-Maze.}
\label{tab:prompt-lmrlmaze}
\centering
\small
\rule{\textwidth}{0.4pt}
\begin{minipage}{0.97\textwidth}
\vspace{0.5em}

\textbf{Prompt:}\\[0.3em]
You are an expert maze solver. Before answering, always:
Step 1. List moves not blocked by walls.
Remember, if there are walls to your right, you can not move right.
If there are walls to your left, you can not move left.
If there are walls above you, you can not move up.
If there are walls below you, you can not move down.
Step 2. Compute the distance for each move.
Step3. Choose the move with the shortest distance and not blocked by walls.

environment: The goal is at position 8, 6. Your current position is at position 1, 5. There are walls to your right, above you, and below you.
Thought:(dx, dy) = (x\_goal - x\_curr, y\_goal - y\_curr) = (8 - 1, 6 - 5) = (7, 1). I am left and up to the goal, move right and move down will make me closer to the goal, but there are walls to my right, above me, and below me, I can NOT move right, move up, move down, since I can only choose to move left.

environment: The goal is at position 8, 6. Your current position is at position 1, 4. There are walls above you, and below you.
Thought:(dx, dy) = (x\_goal - x\_curr, y\_goal - y\_curr) = (8 - 1, 6 - 4) = (7, 2). I am left and up to the goal, but there are walls above me, and below me, since I can NOT move right, move up. Also I have tried move right to position 1, 5, where I can NOT move right, move up, move down, since I can only choose to move left.

environment: The goal is at position 8, 6. Your current position is at position 5, 3. There are walls to your left, below you.
Thought: (dx, dy) = (x\_goal - x\_curr, y\_goal - y\_curr) = (8 - 5, 6 - 3) = (3, 3). I am left and up to the goal, move right and move down will make me closer to the goal, but there are walls to my left, below me, I can NOT move left, move down, since I choose to move right
Action: move right

environment: The goal is at position 8, 6. Your current position is at position 5, 4. There are walls above you, below you.
Thought: (dx, dy) = (x\_goal - x\_curr, y\_goal - y\_curr) = (8 - 5, 6 - 4) = (3, 2). I am left and up to the goal, move right and move down will make me closer to the goal, but there are walls to my left, below me, I can NOT move left, move down, since I choose to move right
Action: move right

environment: The goal is at position 8, 6. Your current position is at position 5, 5. There are walls above you, below you.
Thought: (dx, dy) = (x\_goal - x\_curr, y\_goal - y\_curr) = (8 - 5, 6 - 5) = (3, 1). I am left and up to the goal, move right and move down will make me closer to the goal, but there are walls to my left, below me, I can NOT move left, move down, since I choose to move right
Action: move right

environment: The goal is at position 8, 6. Your current position is at position 5, 6. There is a wall above you.
Thought: (dx, dy) = (x\_goal - x\_curr, y\_goal - y\_curr) = (8 - 5, 6 - 6) = (3, 0). I am up to the goal, move down will make me closer to the goal, there are walls above me, I can NOT move up, since I choose to move down.
Action: move down

environment: The goal is at position 8, 6. Your current position is at position 6, 6. There are walls to your right, to your left.
Thought: (dx, dy) = (x\_goal - x\_curr, y\_goal - y\_curr) = (8 - 6, 6 - 6) = (2, 0). I am up to the goal, move down will make me closer to the goal, there are walls to my right, to my left, I can NOT move left, move right, since I choose to move down.
Action: move down

environment: The goal is at position 8, 6. Your current position is at position 7, 6. There are walls to your right, to your left.
Thought: (dx, dy) = (x\_goal - x\_curr, y\_goal - y\_curr) = (8 - 7, 6 - 6) = (1, 0). I am up to the goal, move down will make me closer to the goal, there are walls to my right, to my left, I can NOT move left, move right, since I choose to move down.
Action: move down

environment: Success

Respond ONLY in the format below without extra text, you can ONLY give one action.

Thought:
\textless{}concise reason\textgreater{}

Action:
\textless{}move up|move down|move left|move right\textgreater{}
\vspace{0.5em}
\end{minipage}
\rule{\textwidth}{0.4pt}
\end{table*}

\subsection{Environment and Dataset Specifications}
We evaluate Fed-SE across five heterogeneous environments. The specifications and prompt designs for each task are derived from the AgentGym suite~\citep{xi2024agentgym}:

\begin{itemize}
    \item \textbf{BabyAI}~\citep{BabyAI} : A grid-world environment for embodied control where agents perform navigation and manipulation tasks.
    \item \textbf{WebShop}~\citep{webshop}: A simulated e-commerce environment for online shopping. The action space includes \texttt{search[]} and \texttt{click[]}.
    \item \textbf{TextCraft}~\citep{textcraft}: A Minecraft-style crafting environment requiring hierarchical planning to craft items using recursive recipes.
    \item \textbf{Maze}~\citep{lmrl}: A long-horizon navigation task testing memory and spatial reasoning capabilities.
    \item \textbf{Wordle}~\citep{lmrl}: A word-guessing game requiring iterative constraint satisfaction and reasoning.
\end{itemize}

For the initial seed data ($\mathcal{D}^{\text{expert}}$), we utilize the \texttt{AgentTraj} dataset provided by AgentGym. The dataset sizes for each task are: WebShop (3930), BabyAI (810), TextCraft (374), Maze (215), and Wordle (955). The evaluation is conducted on test sets, with 200 tasks for WebShop, 100 for TextCraft, 90 for BabyAI, and 25 each for Maze and Wordle..

\subsection{Training Hyperparameters}
We implement Fed-SE using PyTorch and the LLaMA-Factory library. All experiments are conducted on \textbf{4 NVIDIA RTX A6000 GPUs}.

\paragraph{Local Evolution (Client-Side).}
During the local self-evolution phase, we employ Low-Rank Adaptation (LoRA) for parameter-efficient fine-tuning. We freeze the pre-trained backbone and only update the adapter modules injected into all linear layers . For LoRA configuration, we set rank $r=8$, and alpha $\alpha=16$ with target modules set to all linear layers. We use the AdamW optimizer with a learning rate of $5\times 10^{-5}$, cosine learning rate scheduler, and no warmup steps. Each client trains for 2 epochs per communication round. The maximum context length is set to 4096 tokens to accommodate long interaction histories. All training is performed in \texttt{bfloat16} precision for stability and efficiency.

\paragraph{Global Aggregation (Server-Side).}
The federation process spans $T=20$ communication rounds. We establish a setup with $K=5$ clients, where each client is dedicated to one of the five heterogeneous tasks (BabyAI, WebShop, TextCraft, Maze, Wordle). At the end of each round, the server aggregates the LoRA adapters from all clients using unweighted element-wise averaging.

\subsection{Trajectory Collection and Filtering}
\paragraph{Inference Configuration.} 
During the exploration phase, agents interact with the environment using temperature sampling with $T=1.0$ to encourage diverse trajectory generation. The maximum number of interaction turns is limited based on the task difficulty (e.g., 20 rounds for BabyAI, 10 for WebShop).

\paragraph{Filtering Mechanism.}
We implement a strict binary outcome filter. A trajectory $\tau$ is added to the experience buffer $\mathcal{D}^{\text{train}}$ if and only if the environment returns a success flag \texttt{success=1}. Failed trajectories are discarded to prevent negative reinforcement. The experience buffer accumulates data throughout the 20 rounds without a capacity limit.

\subsection{Prompting and Input Construction}
We adopt a \textbf{ReAct (Reason + Act)} prompting strategy. The input to the model consists of:
\begin{enumerate}
    \item \textbf{System Instruction}: Defines the role of the agent, the environment rules, and the valid action format.
    \item \textbf{Interaction History}: The sequence of observations and actions from the current episode.
\end{enumerate}
The agent is required to output a response in the structured format:
\begin{verbatim}
Thought: <Reasoning content>
Action: <Action content>
\end{verbatim}
Prompts are detailed in Tables~\ref{tab:prompt-webshop}--\ref{tab:prompt-lmrlmaze}.

\section{Additional Experiments and Case Study}
\label{sec:case_study}

To provide deeper insights into how Fed-SE improves agent capabilities, we present two categories of analysis: (1) per-environment performance curves across different base models, and (2) qualitative trajectory comparisons demonstrating behavioral changes before and after federated self-evolution.

\subsection{Performance Analysis Across Models}

\begin{figure*}[htbp]
    \centering
    \includegraphics[width=1\linewidth]{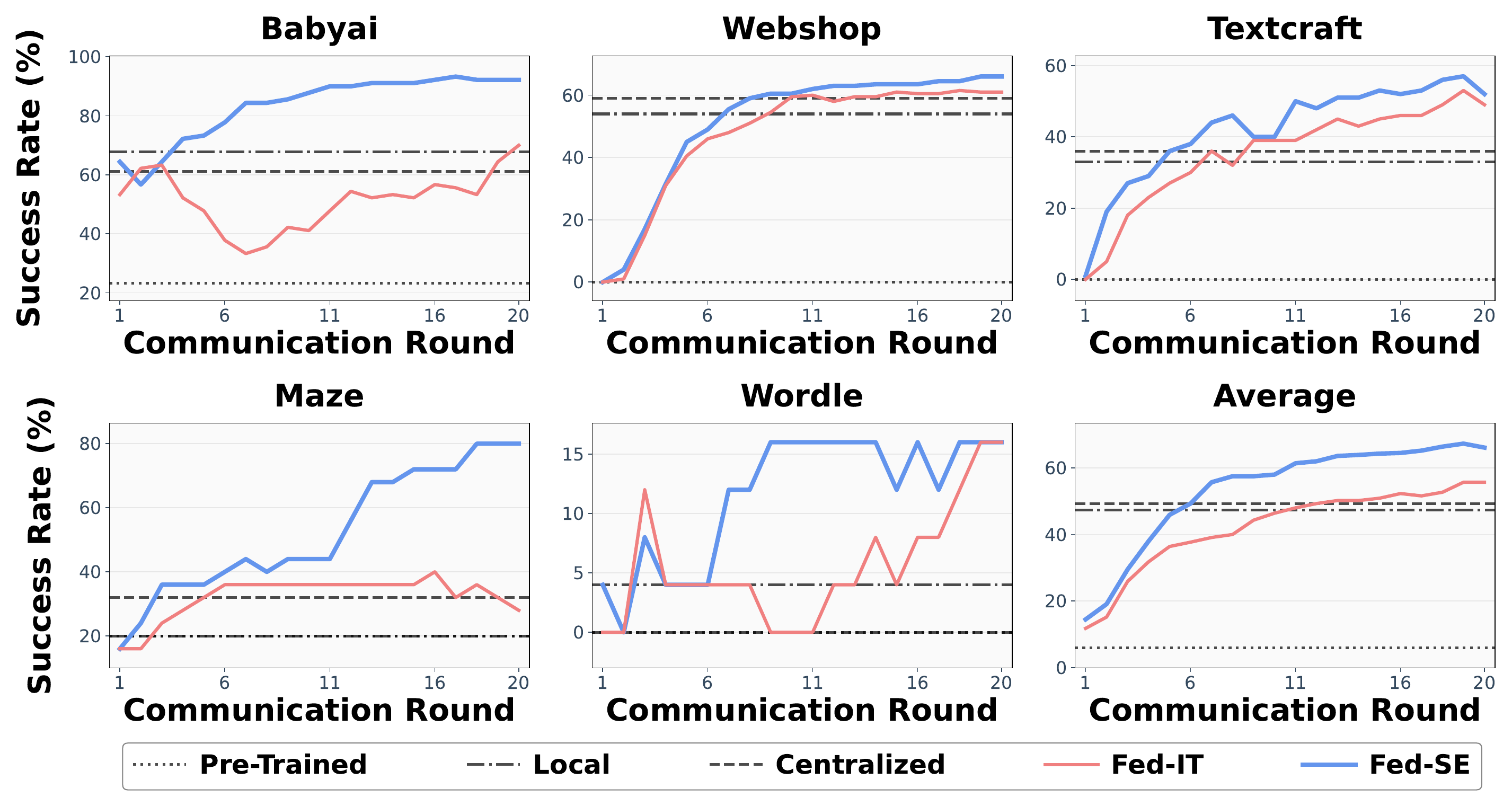}
    \caption{Performance comparison on Llama2-7B across five heterogeneous environments. The curves show task success rates over 20 communication rounds. Fed-SE demonstrates consistent improvements across BabyAI, WebShop, TextCraft, and Maze, achieving performance comparable to or exceeding centralized training while preserving data locality.}
    \label{fig:llama27b}
\end{figure*}

\begin{figure*}[htbp]
    \centering
    \includegraphics[width=1\linewidth]{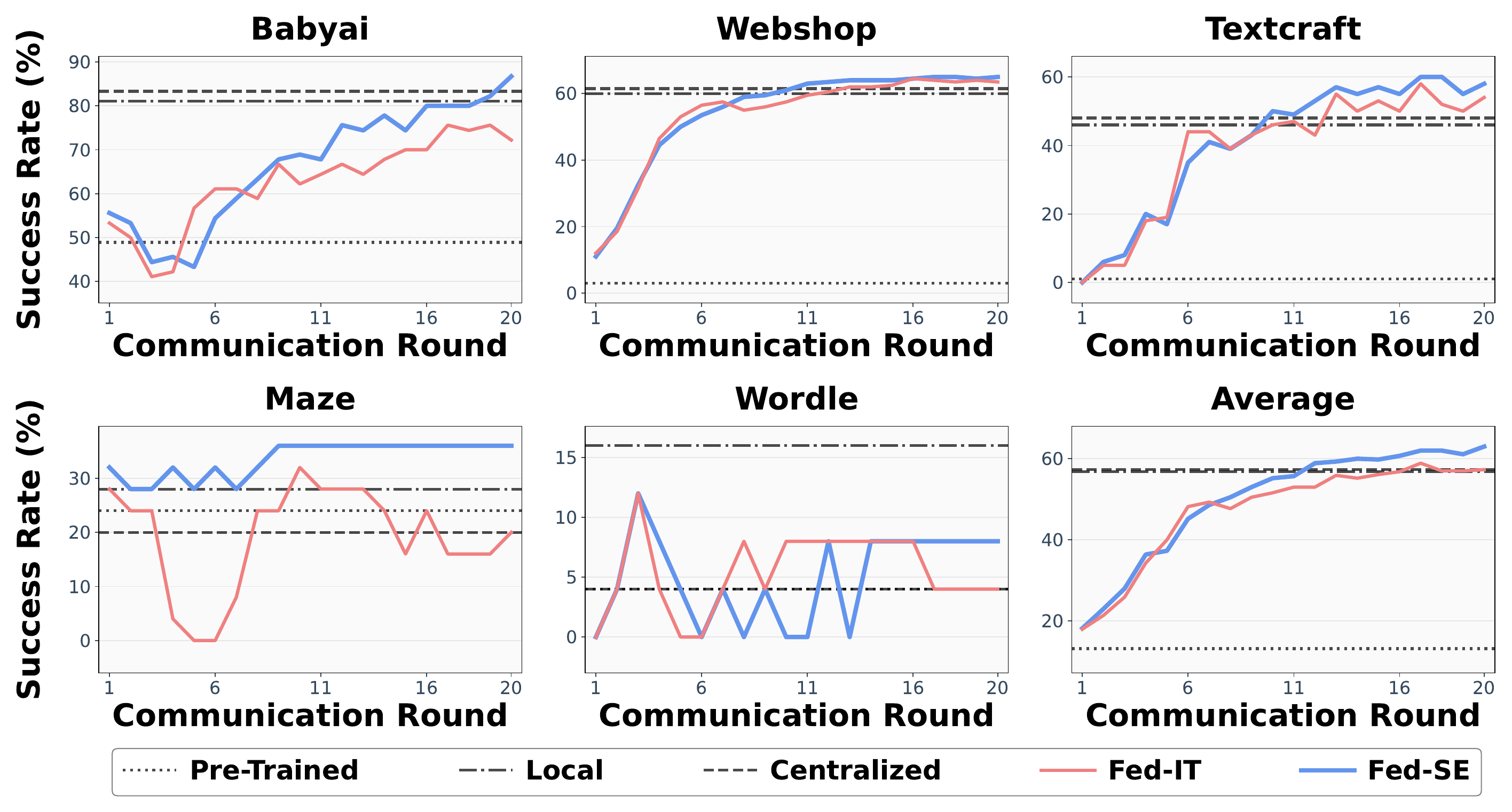}
    \caption{Performance comparison on Qwen2.5-3B. Despite the smaller model capacity, Fed-SE achieves substantial improvements particularly in BabyAI and WebShop environments, demonstrating that the federated self-evolution approach effectively leverages cross-environment knowledge sharing even with limited model parameters.}
    \label{fig:qwen253b}
\end{figure*}

\begin{figure*}[htbp]
    \centering
    \includegraphics[width=1\linewidth]{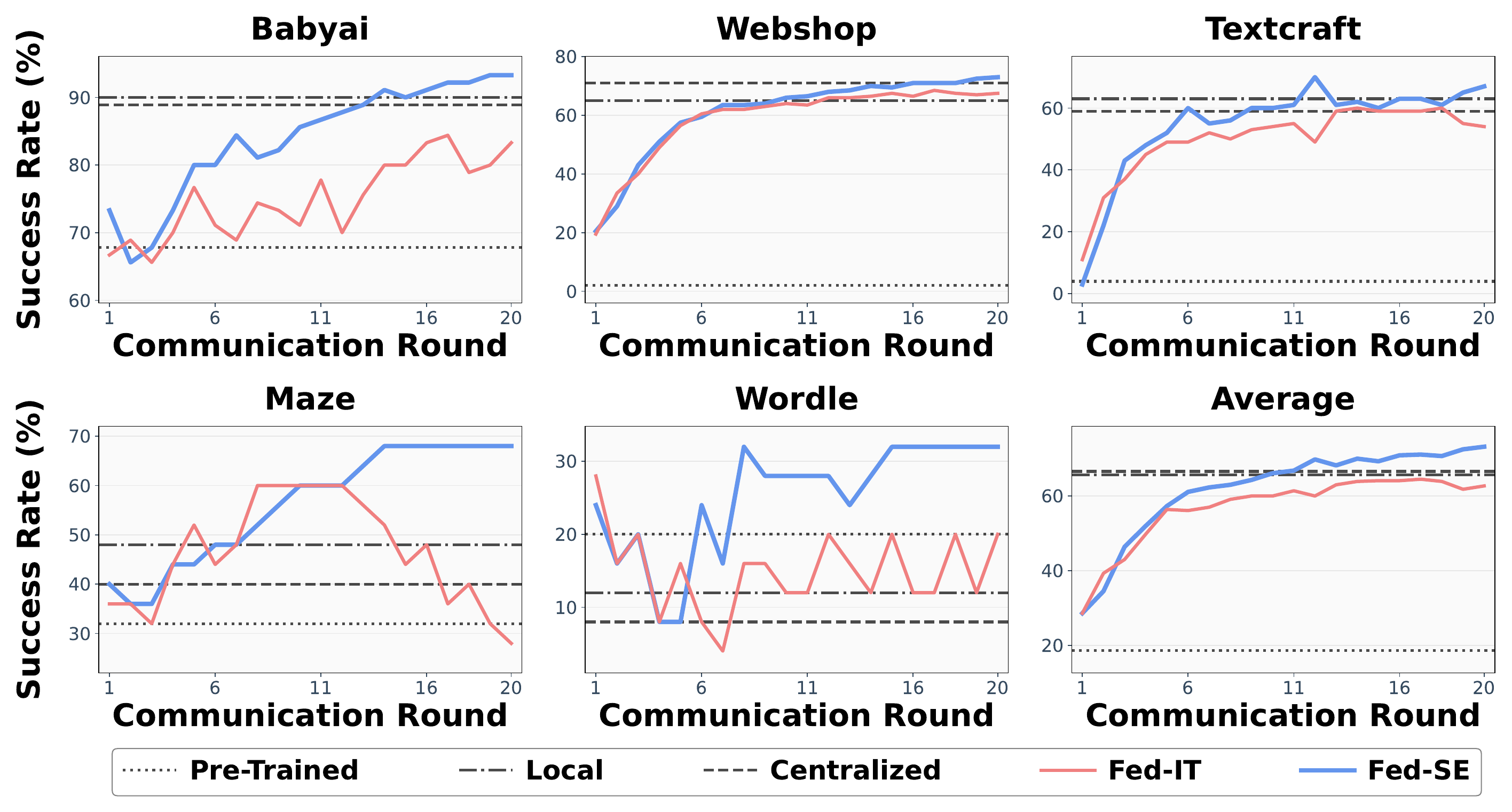}
    \caption{Performance comparison on Qwen2.5-7B. The larger model capacity enables more pronounced improvements across all environments. Fed-SE consistently outperforms Fed-IT and approaches the performance of centralized training, validating the effectiveness of low-rank subspace aggregation in the federated setting.}
    \label{fig:qwen257b}
\end{figure*}

\begin{figure*}[htbp]
    \centering
    \includegraphics[width=1\linewidth]{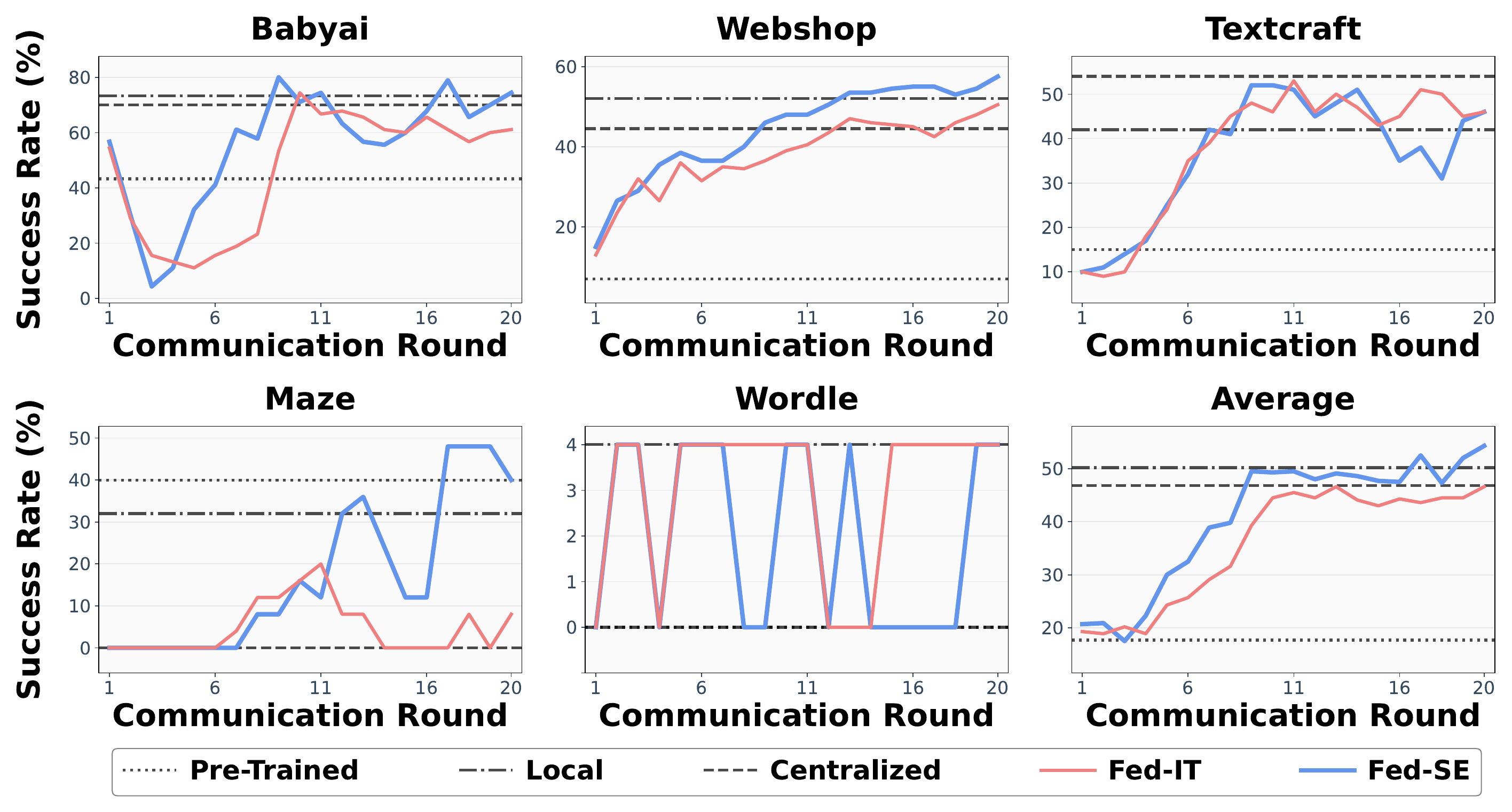}
    \caption{Performance comparison on Qwen3-1.7B. This compact model from the Qwen3 family shows rapid early improvements with Fed-SE, particularly in WebShop and TextCraft environments, demonstrating the sample efficiency of our approach.}
    \label{fig:qwen317b}
\end{figure*}

\begin{figure*}[htbp]
    \centering
    \includegraphics[width=1\linewidth]{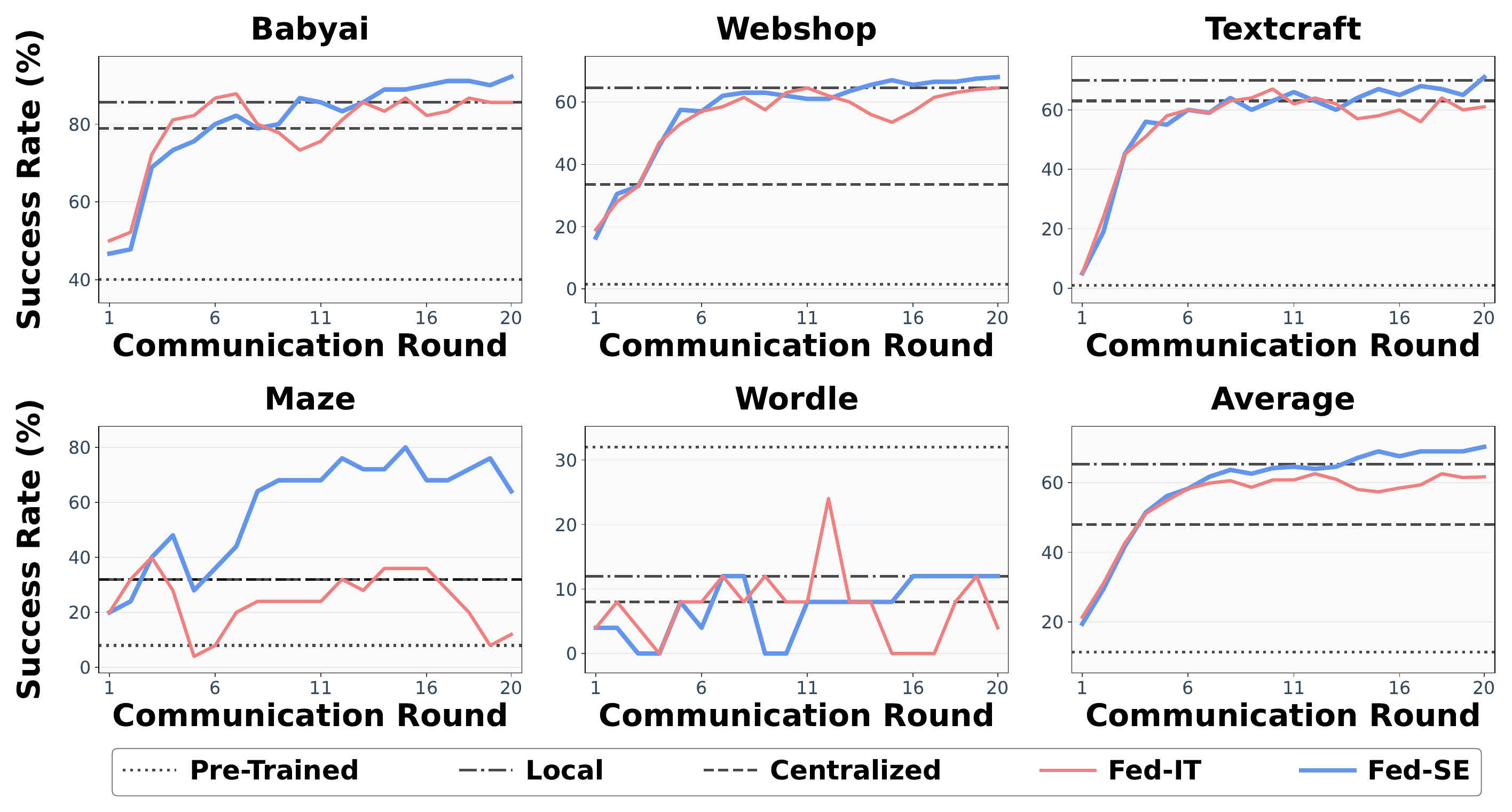}
    \caption{Performance comparison on Qwen3-8B. While this stronger base model achieves higher initial performance, Fed-SE still provides meaningful improvements, though the gains are smaller compared to weaker models due to capability saturation. This aligns with our observation that models with more room for improvement benefit more significantly from federated self-evolution.}
    \label{fig:qwen38b}
\end{figure*}

Figures~\ref{fig:llama27b}--\ref{fig:qwen38b} present the task success rates across five environments for different base models. Each figure shows the performance trajectory over 20 communication rounds, comparing Fed-SE against baseline methods including the pretrained model with no fine-tuning, Local training (without federation), Centralized training (with full data access), and Fed-IT (federated instruction tuning).

\paragraph{Llama2-7B (Figure~\ref{fig:llama27b}).} Fed-SE demonstrates the most substantial improvements on Llama2-7B, achieving 66.1\% average success rate compared to 55.7\% for FedIT. The performance gap is particularly pronounced on Maze, where Fed-SE reaches 80.0\% while FedIT achieves only 28.0\%. On BabyAI, Fed-SE attains 92.2\% success rate, substantially outperforming all baselines. These results indicate that Fed-SE effectively compensates for the relatively weaker instruction-following capabilities of older model architectures.

\paragraph{Qwen2.5-3B (Figure~\ref{fig:qwen253b}).} The smaller-scale Qwen2.5-3B model shows consistent improvements under Fed-SE, reaching 63.0\% average success rate versus 57.3\% for FedIT. Notably, Fed-SE achieves 36.0\% on Maze compared to 20.0\% for FedIT, demonstrating that the framework benefits models with limited parameter capacity. The performance curves reveal stable convergence across all environments, with Fed-SE maintaining advantages throughout the training process.

\paragraph{Qwen2.5-7B (Figure~\ref{fig:qwen257b}).} As the primary evaluation model, Qwen2.5-7B achieves the highest overall performance under Fed-SE with 73.2\% average success rate, surpassing FedIT by 10.5 percentage points. Fed-SE shows improvements across all five environments, with the largest gain observed on Maze where success rate increases from 28.0\% to 68.0\%. The consistent upward trajectory across communication rounds confirms effective knowledge accumulation through online evolution.

\paragraph{Qwen3-1.7B (Figure~\ref{fig:qwen317b}).} Despite its compact size, Qwen3-1.7B under Fed-SE achieves 54.3\% average success rate, outperforming FedIT by 7.7 percentage points. The model maintains competitive performance on Maze, reaching 40.0\% success rate while FedIT drops to 8.0\%. This demonstrates that Fed-SE's trajectory filtering mechanism effectively extracts learning signals even with limited model capacity.

\paragraph{Qwen3-8B (Figure~\ref{fig:qwen38b}).} Fed-SE achieves 70.2\% average success rate on Qwen3-8B, representing an 8.6 percentage point improvement over FedIT. The framework demonstrates particularly strong gains on Maze, improving from 12.0\% under FedIT to 64.0\% under Fed-SE. Interestingly, Wordle performance remains challenging across all methods, suggesting that tasks requiring iterative reasoning with sparse early-stage success signals present persistent difficulties regardless of model scale.

\subsection{Qualitative Trajectory Analysis}

To illustrate the behavioral improvements achieved through Fed-SE, we present representative trajectory comparisons from each environment. Each case demonstrates how the agent's decision-making evolves from suboptimal to successful strategies after federated self-evolution.

\begin{figure*}[htbp]
    \centering
    \includegraphics[width=1\linewidth]{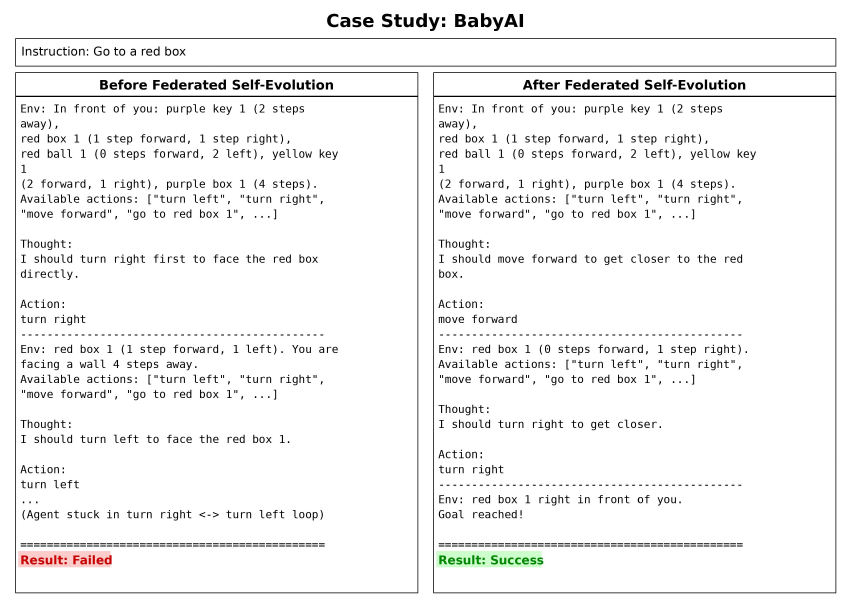}
    \caption{Case study in the BabyAI environment. Task instruction is to navigate to a red box. Before Fed-SE, the agent misinterprets spatial relationships and gets stuck in a turn left/right loop. After Fed-SE, the agent correctly parses position feedback and executes an efficient navigation sequence to reach the goal.}
    \label{fig:case_babyai}
\end{figure*}

\paragraph{BabyAI Environment.}
Figure~\ref{fig:case_babyai} presents a navigation task where the agent must reach a target object (a red box) in a grid world. Before federated self-evolution, the agent fails to effectively parse spatial relationships from environment feedback. When given positional information about surrounding objects, the agent incorrectly interprets the relative positions and enters a repetitive turn left/right loop without making progress toward the goal. After federated self-evolution, the agent demonstrates improved spatial reasoning capabilities. It correctly interprets the position feedback, plans an efficient path by first moving forward to reduce distance, then adjusting direction based on updated environmental observations, ultimately reaching the goal successfully.

\begin{figure*}[htbp]
    \centering
    \includegraphics[width=1\linewidth]{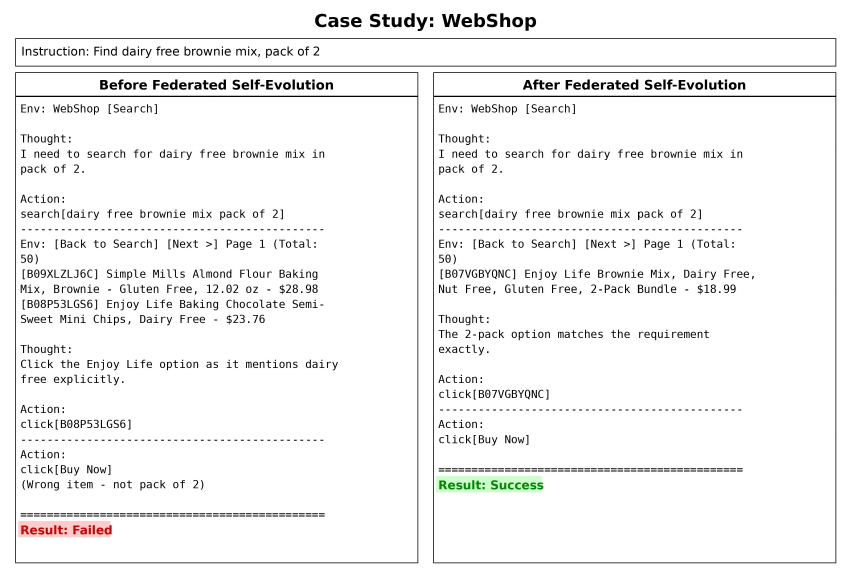}
    \caption{Case study in the WebShop environment. Task instruction is to find dairy free brownie mix in a pack of 2. Before Fed-SE, the agent selects an item without verifying the pack quantity constraint. After Fed-SE, the agent carefully matches product attributes to requirements and selects the correct 2-pack option.}
    \label{fig:case_webshop}
\end{figure*}

\paragraph{WebShop Environment.}
Figure~\ref{fig:case_webshop} demonstrates an online shopping task requiring the agent to find products matching specific constraints. Before federated self-evolution, the agent fails to carefully verify all product attributes against the requirements. In the example shown, the agent selects an item without checking whether it satisfies the pack quantity constraint, resulting in task failure. After federated self-evolution, the agent exhibits more thorough attribute verification behavior. It carefully examines each product's details, identifies that a 2-pack bundle matches the requirement exactly, and completes the purchase successfully. This improvement reflects enhanced constraint satisfaction reasoning acquired through cross-environment knowledge sharing.

\begin{figure*}[htbp]
    \centering
    \includegraphics[width=1\linewidth]{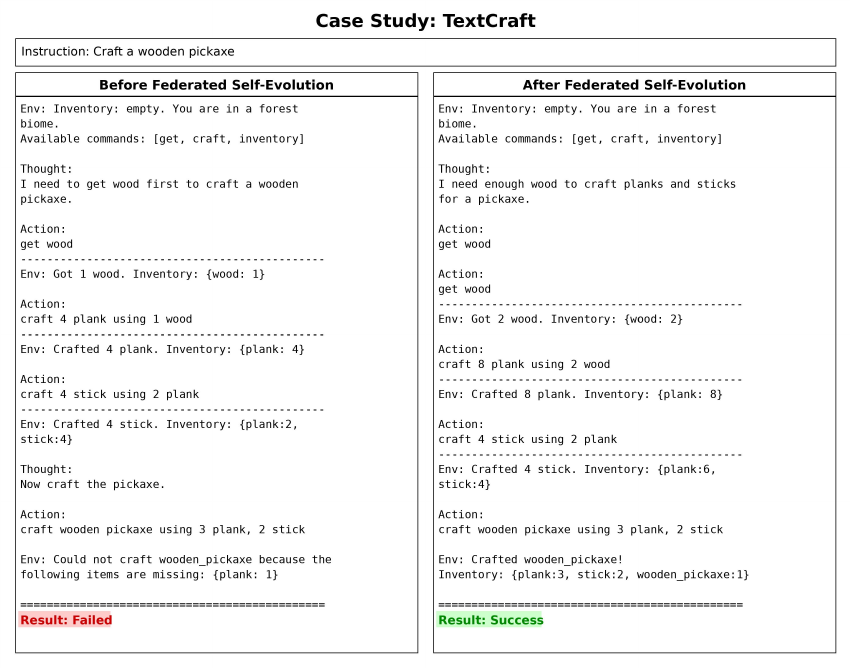}
    \caption{Case study in the TextCraft environment. Task instruction is to craft a wooden pickaxe. Before Fed-SE, the agent gathers insufficient resources (1 wood) and fails at the final crafting step. After Fed-SE, the agent plans ahead by gathering adequate resources (2 wood) and successfully completes the entire crafting sequence.}
    \label{fig:case_textcraft}
\end{figure*}

\paragraph{TextCraft Environment.}
Figure~\ref{fig:case_textcraft} shows a crafting task requiring hierarchical planning to construct a wooden pickaxe. Before federated self-evolution, the agent exhibits insufficient resource gathering behavior. It collects only 1 wood unit and proceeds through the crafting chain, only to fail at the final step due to insufficient planks. After federated self-evolution, the agent demonstrates improved planning capabilities. It anticipates the complete resource requirements by gathering 2 wood units initially, then efficiently progresses through the crafting hierarchy with sufficient materials, successfully completing the task. This improvement highlights the acquisition of hierarchical planning skills through federated experience sharing.

\begin{figure*}[htbp]
    \centering
    \includegraphics[width=1\linewidth]{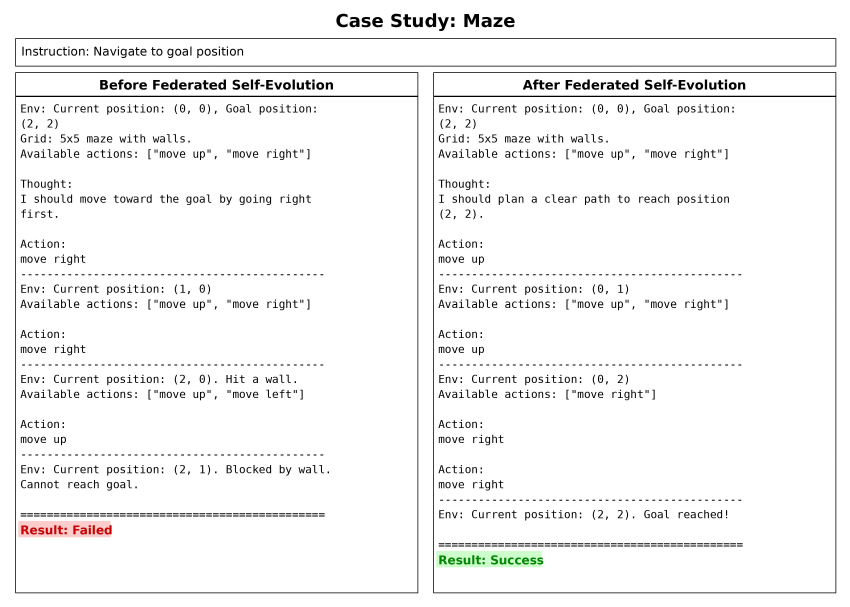}
    \caption{Case study in the Maze environment. Task instruction is to navigate to the goal position (2,2). Before Fed-SE, the agent uses a naive move-until-blocked strategy and gets stuck against walls. After Fed-SE, the agent plans a clear path considering the goal position and navigates efficiently to reach the target.}
    \label{fig:case_maze}
\end{figure*}

\paragraph{Maze Environment.}
Figure~\ref{fig:case_maze} presents a maze navigation task requiring the agent to reach a goal position while avoiding walls. Before federated self-evolution, the agent employs a naive strategy of repeatedly moving in one direction until blocked, then making arbitrary turns, which leads to getting stuck against walls. After federated self-evolution, the agent demonstrates improved path planning capabilities. It considers the goal position when selecting actions, avoids futile movements toward walls, and efficiently navigates to the target using a more strategic approach. This behavioral change reflects enhanced spatial reasoning acquired from cross-environment experiences.

\begin{figure*}[htbp]
    \centering
    \includegraphics[width=1\linewidth]{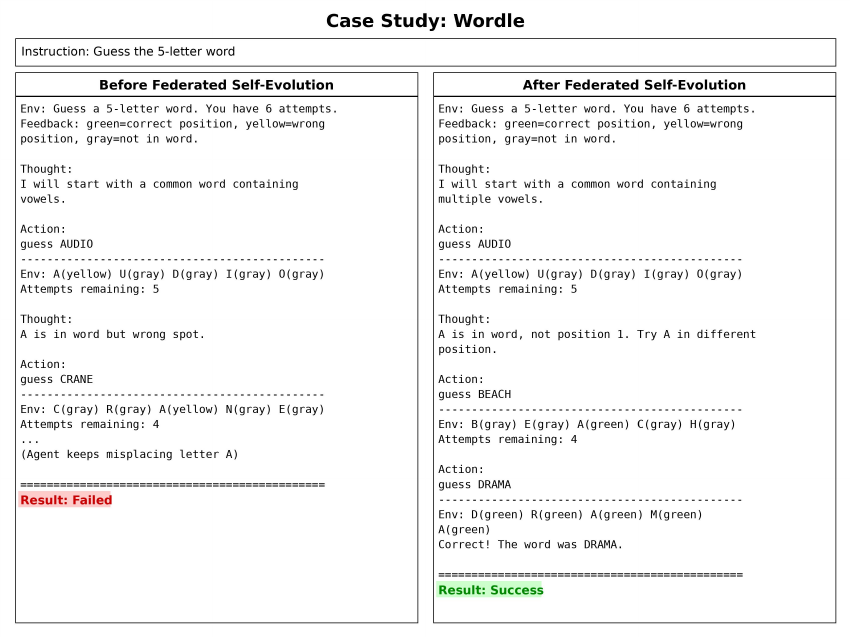}
    \caption{Case study in the Wordle environment. Task instruction is to guess the 5-letter word. Before Fed-SE, the agent ignores positional feedback and keeps misplacing letter ``A''. After Fed-SE, the agent correctly interprets yellow feedback (letter present but wrong position) and strategically repositions letters in subsequent guesses to find the correct word.}
    \label{fig:case_wordle}
\end{figure*}

\paragraph{Wordle Environment.}
Figure~\ref{fig:case_wordle} illustrates a word-guessing task requiring iterative constraint satisfaction. Before federated self-evolution, the agent fails to properly utilize the feedback from previous guesses. When receiving feedback that letter ``A'' is in the word but in the wrong position (yellow), the agent repeatedly guesses words with ``A'' in similar positions, never successfully repositioning the letter. After federated self-evolution, the agent demonstrates improved constraint reasoning. Upon receiving yellow feedback for ``A'', it explicitly reasons about trying ``A'' in a different position and selects subsequent guesses accordingly, ultimately finding the correct word. This improvement reflects enhanced iterative reasoning capabilities.

\end{document}